\documentclass{article} 
\usepackage{iclr2027_conference,times}

\usepackage{amsmath,amsfonts,bm}

\def\eqref#1{equation~\ref{#1}}

\def\1{\bm{1}}

\DeclareMathAlphabet{\mathsfit}{\encodingdefault}{\sfdefault}{m}{sl}
\SetMathAlphabet{\mathsfit}{bold}{\encodingdefault}{\sfdefault}{bx}{n}

\usepackage{hyperref}
\usepackage{url}

\usepackage{multirow}    
\usepackage{xcolor}      
\usepackage{ulem}        
\usepackage{booktabs}    
\usepackage{makecell}    
\usepackage{caption}     
\usepackage{rotating}  

\usepackage{enumitem}   
\usepackage{amsmath}    

\usepackage{wrapfig}

\usepackage{graphicx}  
\usepackage{float}     

\definecolor{second_blue}{RGB}{0, 0, 255}    
\definecolor{first_red}{RGB}{255, 0, 0}    
\definecolor{year_green}{RGB}{0, 150, 0}    

\title{RouteTS: Frequency-Time Routing for Time Series Forecasting}

\author{
  Gaofeng Lin, Lei Duan\thanks{Corresponding author.} \\
  School of Computer Science, Sichuan University \\
  \texttt{gaofenglin@stu.scu.edu.cn}, \texttt{leiduan@scu.edu.cn} \\
}

\iclrfinalcopy 
\begin{document}

\maketitle

\begin{abstract}
Real-world time series inherently intertwine global periodic structures with localized non-stationary variations. Existing approaches process these heterogeneous dynamics within a single computational domain, incurring fundamental limitations: time-domain models suffer from periodic misalignment over long horizons, while frequency-domain models over-smooth transient spikes. We argue that the optimal computational domain is not a property of the model, but of the data itself. Based on this principle, we propose RouteTS, a unified forecasting framework that partitions the frequency spectrum via amplitude routing and delegates components to their mathematically optimal domains. Dominant frequencies are processed by a complex-valued linear predictor in the frequency domain to preserve periodic structure, while residual spectral energy is reverted to the time domain and modeled by a lightweight MLP for local variations. Extensive experiments demonstrate that RouteTS achieves competitive prediction accuracy across diverse real-world datasets, with routing decisions guided by the underlying spectral signature. Furthermore, the lightweight design of RouteTS provides significant computational efficiency advantages, offering a principled solution to the longstanding dilemma between global periodicity and local transience.
\end{abstract}

\section{INTRODUCTION}
\label{sec:intro}

Time series forecasting (TSF) is crucial for applications such as energy planning\citep{chen2011energy, lai2018modeling, ju2021novel}, financial markets \citep{He2022InstanceBased} and traffic management\citep{cirstea2021enhancenet, wu2022autocts}. Real-world time series are characterized by complex temporal dynamics that inherently intertwine global structural patterns with localized non-stationary variations. Effectively modeling these heterogeneous components is fundamental to robust forecasting.

\begin{figure}[t]
\centering
\includegraphics{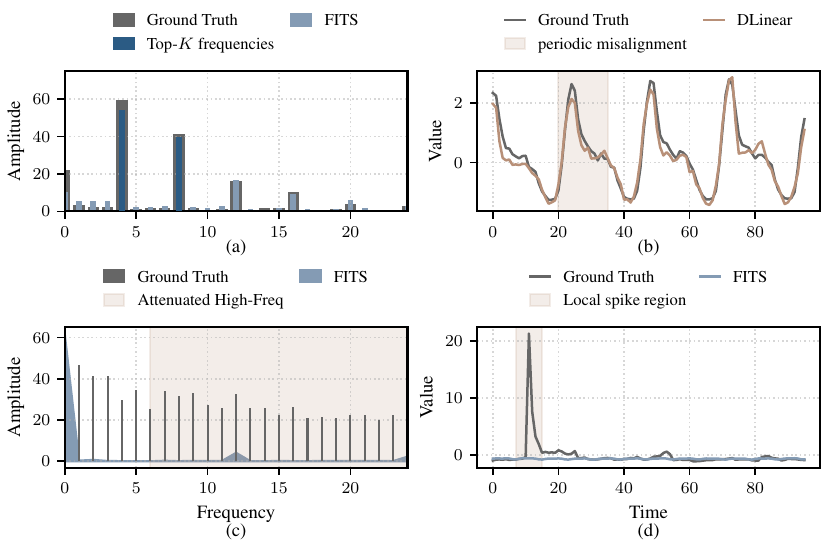}
\caption{Illustration of the dilemma in single-domain forecasting models. (a) and (b) visualize periodic-dominant data (Traffic): (a) the sparse frequency spectrum with energy concentrated in Top-K frequencies; (b) time-domain point-to-point mapping suffers from gradual phase misalignment. (c) and (d) visualize transient-dominant data (PM2.5): (c) the frequency spectrum where the shaded area indicates severe attenuation of high-frequency components; (d) FITS reconstruction severely over-smooths localized transient spikes that are present in the Ground Truth.}
\label{fig:single-domain_dilemma}
\end{figure}

Recently, deep learning models have achieved encouraging results in time series forecasting. However, most existing architectures process complex temporal dynamics within a single computational domain, incurring inherent limitations. For periodic-dominant signals, the frequency spectrum exhibits sparse high-amplitude peaks concentrated in a few frequency components, as shown in Figure~\ref{fig:single-domain_dilemma}(a), indicating that frequency-domain modeling is well-suited to capture the global periodic structure. Nevertheless, simple time-domain models that learn point-to-point mappings inherently lack explicit periodic constraints. Without built-in geometric priors for periodic structures, these models must implicitly learn cycle alignment from data, making them susceptible to accumulated local errors over long horizons. As illustrated in Figure~\ref{fig:single-domain_dilemma}(b), this manifests as periodic misalignment, where the predicted cycles deviate from the ground truth in both timing and amplitude.

Conversely, real-world time series are not merely ideal superpositions of periodic patterns. For transient-dominant data, non-stationary variations and sharp spikes are localized in time. When transformed to the frequency domain, these localized features spread across numerous frequency components, as shown in Figure~\ref{fig:single-domain_dilemma}(c). Yet compact frequency representations such as FITS retain only a limited subset of dominant components, inevitably leading to over-smoothing of sharp edges in the time-domain reconstruction, as illustrated in Figure~\ref{fig:single-domain_dilemma}(d). In sharp contrast, direct time-domain processing can precisely capture such local non-linear spikes. This fundamental dilemma reveals that no single domain is universally optimal.

We argue that the answer lies in the spectral structure of time series itself. Global periodicity, which dominates certain datasets, manifests as sparse high-amplitude spectral components; conversely, localized transient variations and non-stationary fluctuations, prevalent in others, are characterized by a diffuse low-amplitude spectral floor and are best preserved through direct time-domain processing. This complementary spectral–temporal structure suggests a natural domain-routing strategy: dominant frequency components capturing global structure should be processed in the frequency domain, while the residual spectral energy, which encodes local variations, is delegated to the time domain.

Motivated by these observations, we propose RouteTS, a unified forecasting framework that partitions the frequency spectrum and routes its components to their mathematically optimal domains. Concretely, for a given time series, we first transform it to the frequency domain and identify the Top-$K$ dominant frequencies. These high-amplitude components are processed by a complex-valued linear predictor that preserves their inherent periodic structure, while the remaining low-amplitude components are mapped back to the time domain via inverse FFT and handled by a lightweight MLP. The final prediction is obtained by summing both branches. By aligning the physical characteristics of the data with their optimal computational domains, our approach sidesteps both time-domain phase misalignment and frequency-domain over-smoothing, bridging the gap between global periodic structure and local transient dynamics. The key insight of RouteTS is that the optimal computational domain is not a property of the model, but of the data itself—and can be determined from its spectral structure.
\section{RELATED WORK}
\subsection{Time-Domain Deep Forecasting}

Early RNN-\citep{ma_lstm_2015,rangapuram_deep_2018,salinas_deepar_2020} and TCN-based\citep{bai_empirical_2018,liu_time_2021} models laid the groundwork for time series forecasting, but their limited long-range capacity has shifted focus toward Transformer and Linear architectures.

Transformer-based methods leverage self-attention for global dependency modeling, evolving from per-step tokens \citep{zhou2021informer}, to patched subsequences \citep{wu2021autoformer,nie2023time}, to inverted variate tokens \citep{liu2024itransformer}, and recently to phase-aligned representations \citep{niu2026phaseformer}. Despite their expressive power, these models lack explicit periodic constraints, forcing implicit cycle learning through high-dimensional parameterization that renders them computationally heavy and prone to overfitting non-stationary noise.

Linear-based methods, epitomized by DLinear \citep{zeng_are_2023}, have shown that simple linear mappings with trend-seasonal decomposition often outperform sophisticated Transformers. While multi-scale feature fusion has advanced this line \citep{hu_adaptive_2025}, compressing all temporal dynamics into a single linear projection inevitably constrains capacity, rendering these approaches sub-optimal for highly non-stationary abrupt changes.

Both paradigms share a critical limitation: entangling deterministic periodicities and stochastic high-frequency variations within a unified space forces a dilemma—either overfitting to local noise or blurring genuine periodic signals—motivating us to transcend single-domain processing and route heterogeneous dynamics to their optimal domains.

\subsection{Frequency-Domain Forecasting}

Recent frequency-domain methods transform time series via FFT, manipulate spectral features, and reconstruct predictions through IFFT. Pure frequency models such as FITS \citep{xu_fits_2024} and FilterNet \citep{yi_filternet_2024} struggle with localized transients: non-stationary spikes spread across numerous frequency components, and sparse spectral representations inevitably over-smooth sharp edges. Subsequent works incorporate time-domain information to mitigate this. FBM \citep{runze2024rethinking} constructs joint time-frequency features via orthogonal basis functions, while CFPT \citep{kou_cfpt_2025} and FilterTS \citep{wang_filterts_2025} employ mode decomposition and adaptive frequency filtering. MixLinear \citep{ma2026mixlinear} further combines segment-based trend extraction with low-rank spectral filtering.

However, these approaches rely on static domain allocation. Pure frequency models force all dynamics into the complex spectral space, where retaining limited components causes over-smoothing while including more high-frequency components amplifies noise. Joint and dual-pathway models impose fixed structural priors: MixLinear rigidly assigns local trends to the time domain and global trends to the frequency domain regardless of dataset characteristics. Such static separation precludes dynamic adaptation, leaving existing methods unable to balance deterministic periodicities and stochastic residuals across heterogeneous datasets.

\section{Method}

\begin{figure}[htbp]
\begin{center}
\includegraphics[width=1\textwidth]{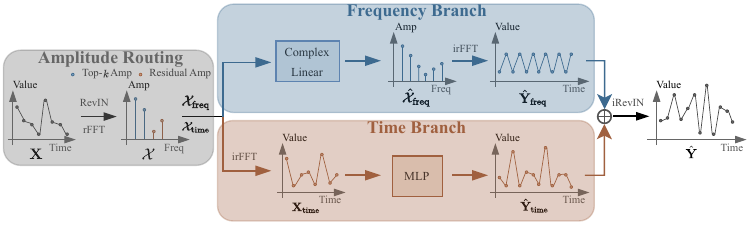}
\end{center}
\caption{Overall architecture of RouteTS. The input spectrum is partitioned via an amplitude routing mechanism: dominant frequency components are routed to the Frequency Branch for complex-valued periodic forecasting, while the remaining residual spectrum is reverted to the time domain. This allows the Time Branch to effectively capture non-stationary local variations via an MLP backbone before the final time-domain aggregation.}
\label{fig:architecture}
\end{figure}

In this section, we present the details of our frequency-time routing architecture. Section~\ref{subsec:amplitude_routing} introduces the amplitude-based routing mechanism used to split the input signal. Section~\ref{subsec:frequency_branch} and Section~\ref{subsec:time_branch} subsequently detail the specific operations of the Frequency-Domain main predictor in the frequency domain and the Time-Domain residual predictor in the time domain. The final prediction is elegantly formulated as an additive aggregation of the two predictors. An overview of the entire data flow is illustrated in Figure~\ref{fig:architecture}.

\noindent\textbf{Task Formulation.} We consider a regularly sampled multivariate time series comprising $C$ distinct variables. Given a current timestamp $t$, the model takes a historical lookback window of length $L$ as input, directly denoted as $\mathbf{X} = [\mathbf{x}_{t-L+1}, \mathbf{x}_{t-L+2}, \dots, \mathbf{x}_t] \in \mathbb{R}^{L \times C}$, where $\mathbf{x}_i \in \mathbb{R}^{C}$ represents the observations at step $i$. Correspondingly, the forecasting target is the future horizon window of length $T$, defined as $\mathbf{Y} = [\mathbf{x}_{t+1}, \mathbf{x}_{t+2}, \dots, \mathbf{x}_{t+T}] \in \mathbb{R}^{T \times C}$. The core objective is to learn a mapping function that accurately predicts the future values $\hat{\mathbf{Y}}$ based solely on the input sequence $\mathbf{X}$.

\subsection{Amplitude Routing}
\label{subsec:amplitude_routing}

To mitigate non-stationary distribution shifts in real-world time series, we first apply Reversible Instance Normalization (RevIN) to the input sequence $\mathbf{X} \in \mathbb{R}^{L \times C}$, yielding the normalized sequence $\mathbf{X}_{\text{norm}}$. We then apply real Fast Fourier Transform (rFFT) to map the signal to frequency domain:

$$\mathcal{X} = \text{rFFT}(\mathbf{X}_{\text{norm}}) \in \mathbb{C}^{F \times C}$$
where $F = \lfloor L/2 \rfloor + 1$ denotes the number of retained frequency components.

Next, we compute the amplitude spectrum $\mathcal{A} = |\mathcal{X}| \in \mathbb{R}^{F \times C}$. For each variable, we select the Top-$K$ dominant frequencies by amplitude and construct a binary routing mask $\mathcal{M} \in \{0, 1\}^{F \times C}$:
$$\mathcal{M}_{f, c} = \begin{cases}
1, & \text{if } \mathcal{A}_{f, c} \in \text{Top-}K(\mathcal{A}_{:, c}) \\
0, & \text{otherwise}
\end{cases}$$
where $K$ is a hyperparameter controlling the number of dominant frequency components allocated to the Frequency Branch. Using this mask, the spectrum is partitioned into two complementary streams:
$$\mathcal{X}_{\text{freq}} = \mathcal{X} \odot \mathcal{M}, \quad \mathcal{X}_{\text{time}} = \mathcal{X} \odot (1 - \mathcal{M})$$
where $\mathcal{X}_{\text{freq}}$ captures the global deterministic periodicities routed to the Frequency Branch, and $\mathcal{X}_{\text{time}}$ retains the non-stationary local variations routed to the Time Branch.

\subsection{Frequency Branch}
\label{subsec:frequency_branch}

After isolating the principal frequency components $\mathcal{X}_{\text{freq}}$, the goal of the Frequency Branch is to forecast the future spectrum $\hat{\mathcal{X}}_{\text{freq}} \in \mathbb{C}^{F' \times C}$, where $F' = \lfloor T/2 \rfloor + 1$ corresponds to the frequency dimension of the prediction horizon $T$.

To preserve the inherent coupling between amplitude and phase in the frequency domain, we formulate the frequency-domain forecasting as a direct complex-valued linear projection. Let $\mathcal{X}_{\text{freq}}$ be decomposed into its real and imaginary parts: $\mathcal{X}_{\text{freq}} = \mathcal{X}_{R} + i\mathcal{X}_{I}$. Adopting the Channel Independence strategy, we introduce a shared learnable complex weight matrix $\mathbf{W} = \mathbf{W}_{R} + i\mathbf{W}_{I} \in \mathbb{C}^{F \times F'}$ applied identically across all $C$ variables, which maps the historical frequency dimension $F$ directly to the future frequency dimension $F'$.

Following the algebraic rules of complex multiplication, the future spectrum is computed via two parallel real-valued linear layers parameterized by $\mathbf{W}_{R}$ and $\mathbf{W}_{I}$:$$\begin{aligned}
\hat{\mathcal{X}}_{R} &= \mathcal{X}_{R} \mathbf{W}_{R} - \mathcal{X}_{I} \mathbf{W}_{I} \\
\hat{\mathcal{X}}_{I} &= \mathcal{X}_{R} \mathbf{W}_{I} + \mathcal{X}_{I} \mathbf{W}_{R}
\end{aligned}$$

where $\hat{\mathcal{X}}_{R}, \hat{\mathcal{X}}_{I} \in \mathbb{R}^{F' \times C}$ are the predicted real and imaginary components, respectively. This formulation explicitly learns both amplitude scaling and phase shifting without relying on intermediate inverse transformations.

Finally, the predicted complex spectrum $\hat{\mathcal{X}}_{\text{freq}} = \hat{\mathcal{X}}_{R} + i\hat{\mathcal{X}}_{I}$ is transformed back to the time domain via the inverse real Fast Fourier Transform (irFFT), yielding the time-domain principal cycle prediction:
$$\hat{\mathbf{Y}}_{\text{freq}} = \text{irFFT}(\hat{\mathcal{X}}_{\text{freq}}) \in \mathbb{R}^{T \times C}$$

\subsection{Time Branch}
\label{subsec:time_branch}

While the principal frequencies successfully capture global periodicities, the residual spectrum $\mathcal{X}_{\text{time}}$ predominantly encapsulates low-amplitude components corresponding to non-stationary local variations. Modeling such abrupt dynamics directly in the frequency domain is inefficient and prone to artifacts. Thus, we revert the residual spectrum back to the time domain via the inverse real Fast Fourier Transform (irFFT):

$$\mathbf{X}_{\text{time}} = \text{irFFT}(\mathcal{X}_{\text{time}}) \in \mathbb{R}^{L \times C}$$

To forecast the future residual variations $\hat{\mathbf{Y}}_{\text{time}} \in \mathbb{R}^{T \times C}$, we employ a MLP backbone that captures complex non-linear local dependencies:

$$\hat{\mathbf{Y}}_{\text{time}} = \text{Linear}(\text{Dropout}(\text{GELU}(\text{Linear}(\mathbf{X}_{\text{time}}))))$$

Final Aggregation. Finally, the predictions from both branches are directly aggregated in the time domain. An inverse RevIN (iRevIN) operation is subsequently applied to restore the original data scale and distribution, yielding the ultimate prediction:

$$\hat{\mathbf{Y}} = \text{iRevIN}(\hat{\mathbf{Y}}_{\text{freq}} + \hat{\mathbf{Y}}_{\text{time}})$$

This explicit divide-and-conquer paradigm ensures that the distinct physical properties of stable structural cycles and non-stationary local residuals are optimally preserved and forecast.

\section{Experiments}

\subsection{Setup}
\label{subsec:setup}

\noindent\textbf{Datasets.}  Our evaluations are conducted on eight real-world multivariate time series datasets spanning diverse domains, including energy, transportation, economics, environment, and healthcare. Specifically, we adopt ILI, COVID-19, PM2.5, Traffic, Solar, ETTh2, ETTm2 and Exchange datasets, following standard time series forecasting literature. More comprehensive details regarding dataset statistics are provided in Appendix~\ref{app:datasets}.

\noindent\textbf{Baselines.} To rigorously evaluate the effectiveness of our RouteTS, we compare it against ten well-acknowledged forecasting models, including PhaseFormer\citep{niu2026phaseformer}, MixLinear\citep{ma2026mixlinear}, CFPT\citep{kou_cfpt_2025}, FilterTS\citep{wang_filterts_2025}, AMD\citep{hu_adaptive_2025}, FBM\citep{runze2024rethinking}, iTransformer\citep{liu2024itransformer}, FilterNet\citep{yi_filternet_2024}, FITS\citep{xu_fits_2024} and DLinear\citep{zeng_are_2023}. These baselines encompass a wide spectrum of modeling paradigms, including advanced Transformer-based architectures, streamlined linear mappings, and recent frequency/decoupling-based networks. Additional details about the baseline configurations are available in Appendix~\ref{app:baselines}.

\noindent\textbf{Implementation Details.}  Our model is implemented with PyTorch1.11\citep{paszke_pytorch_}, trained using the Adam\citep{Adam} optimizer, and all experiments are conducted on a single NVIDIA RTX 3090 GPU (24GB). We adopt MSE (Mean Squared Error) as the loss function and report both MSE and MAE (Mean Absolute Error) as the evaluation metrics. For exhaustive implementation details, please refer to Table~\ref{tab:hyperparams} for the specific routing threshold $K$, and Appendix~\ref{app:implementation} for other dataset-specific hyperparameter configurations.

\begin{table}[htbp]
\centering
\footnotesize
\caption{Dataset-specific hyperparameter configurations for RouteTS. $K$ denotes the number of dominant frequencies routed to the Frequency Branch; $K=0$ indicates pure time-domain processing, $K=\text{full}$ indicates pure frequency-domain processing, and intermediate values indicate hybrid routing.}
\renewcommand{\arraystretch}{1.1}
\setlength{\tabcolsep}{6pt}
\begin{tabular}{c | cccccccc}
\toprule
Datasets & ILI & COVID-19 & PM2.5 & Traffic & Solar & ETTh2 & ETTm2 & Exchange \\
\midrule
$K$ & 0 & 0 & 0 & 1 & 1 & full & full & full \\
\bottomrule
\end{tabular}
\label{tab:hyperparams}
\end{table}

The remainder of this section evaluates RouteTS from three complementary perspectives. We first benchmark forecasting accuracy against  baselines to establish quantitative competitiveness. We then analyze the spectral mechanisms and visualize internal routing behaviors to explain why the proposed decomposition is effective. Finally, we assess memory footprint and parameter efficiency to validate the practical deployability of the framework .

\subsection{Main Results}
\label{sec:main_results}

Table~\ref{tab:main_results} summarizes the multivariate forecasting performance. RouteTS employs dataset-specific routing thresholds $K$ (Table~\ref{tab:hyperparams}), reflecting the principle that the optimal computational domain is determined by the data's spectral structure rather than being fixed a priori.

Overall, RouteTS achieves the best average MSE on six of the eight benchmarks and maintains competitive MAE across most datasets. This validates the core premise that routing heterogeneous temporal dynamics to their mathematically optimal domains yields robust forecasting. On the highly non-stationary COVID-19 dataset, RouteTS secures the lowest MSE and MAE, effectively capturing abrupt localized spikes through pure time-domain processing. On Solar, RouteTS achieves the best results in both metrics, demonstrating that the dual-branch collaboration successfully balances dominant periodicities and residual local variations.

The Traffic dataset presents a challenging scenario where iTransformer achieves the best MSE and PhaseFormer the best MAE. This aligns with the observation that Traffic exhibits strong spatiotemporal dependencies across its 862 channels, where the flow at one detection point significantly influences neighboring points. iTransformer benefits from this structure through its inverted attention mechanism that models inter-channel relationships. PhaseFormer achieves strong performance through its explicit phase-space tokenization, which precisely characterizes the sharp periodic structure and intra-cycle fluctuations prevalent in Traffic. RouteTS, following a channel-independent design, processes each variable separately and thus does not exploit these cross-sensor correlations, though it still maintains competitive performance.

Similarly, on the Exchange dataset, RouteTS trails lightweight linear baselines such as DLinear. Financial exchange rates exhibit weak periodic structure and are dominated by high-frequency stochastic fluctuations, where aggressive smoothing from simple linear mappings can act as an implicit noise suppressor. Nevertheless, RouteTS maintains competitive robustness within a single unified framework, adapting its routing strategy without architectural modification.

\begin{table}[htbp]
\centering
\scriptsize  
\setlength{\tabcolsep}{1.5pt}  
\renewcommand{\arraystretch}{1.2}  
\resizebox{\linewidth}{!}{%
\begin{tabular}{c | cc | cc | cc | cc | cc | cc | cc | cc | cc | cc | cc}
\toprule
\textbf{Model}
& \multicolumn{2}{c|}{\makecell{\textbf{RouteTS}\\\textbf{(our)}}}
& \multicolumn{2}{c|}{\makecell{PhaseFormer\\(2026)}}
& \multicolumn{2}{c|}{\makecell{MixLinear\\(2026)}}
& \multicolumn{2}{c|}{\makecell{CFPT\\(2025)}}
& \multicolumn{2}{c|}{\makecell{FilterTS\\(2025)}}
& \multicolumn{2}{c|}{\makecell{AMD\\(2025)}}
& \multicolumn{2}{c|}{\makecell{FBM\\(2024)}}
& \multicolumn{2}{c|}{\makecell{iTransformer\\(2024)}}
& \multicolumn{2}{c|}{\makecell{FilterNet\\(2024)}}
& \multicolumn{2}{c|}{\makecell{FITS\\(2024)}}
& \multicolumn{2}{c}{\makecell{DLinear\\(2023)}} \\
\midrule
\textbf{Metric}
& MSE & MAE & MSE & MAE & MSE & MAE & MSE & MAE & MSE & MAE
& MSE & MAE & MSE & MAE & MSE & MAE & MSE & MAE & MSE & MAE
& MSE & MAE \\
\midrule
ILI        & \textcolor{red}{\textbf{1.640}} & \textcolor{red}{\textbf{0.807}} & 2.726 & 1.114 & 2.516 & 1.058 & 2.144 & \underline{\textcolor{second_blue}{0.933}} & 2.132 & 0.936 & 3.085 & 1.261 & \underline{\textcolor{second_blue}{2.205}} &0.956 & 2.579 & 1.075 & 2.232 & \underline{\textcolor{second_blue}{0.933}} & 2.900 & 1.190 & 2.427 & 1.047 \\
\midrule
COVID-19   & \textcolor{red}{\textbf{0.125}} & \textcolor{red}{\textbf{0.135}} & 0.211 & \underline{\textcolor{second_blue}{0.192}} & 1.188 & 0.470 & 0.416 & 0.233 & 0.310 & 0.260 & 1.336 & 0.568 &1.219 &0.363 & 0.943 & 0.431 & 0.271 & 0.234 & 0.764 & 0.384 & \underline{\textcolor{second_blue}{0.182}} & 0.283 \\
\midrule
PM2.5      & \textcolor{red}{\textbf{0.416}} & \textcolor{red}{\textbf{0.424}} & 0.427 & 0.429 & 0.426 & 0.438 & 0.419 & 0.428 & 0.421 & 0.427 & 0.436 & 0.431 &0.440 &0.430 & \underline{\textcolor{second_blue}{0.420}} & \underline{\textcolor{second_blue}{0.427}} & 0.437 & 0.440 & 0.422 & 0.432 & 0.420 & 0.467 \\
\midrule
Traffic    & 0.402 & \underline{\textcolor{second_blue}{0.274}} & \underline{\textcolor{second_blue}{0.397}} & \textcolor{red}{\textbf{0.258}} & 0.441 & 0.294 & 0.410 & 0.274 & 0.411 & 0.286 & 0.420 & 0.289 & 0.405 & 0.276 & \textcolor{red}{\textbf{0.386}} & \underline{\textcolor{second_blue}{0.274}} & 0.403 & 0.282 & 0.434 & 0.291 & 0.454 & 0.328 \\
\midrule
Solar      & \textcolor{red}{\textbf{0.211}} & \underline{\textcolor{second_blue}{0.258}} & \underline{\textcolor{second_blue}{0.217}} & \textcolor{red}{\textbf{0.252}} & 0.290 & 0.298 & 0.221 & 0.270 & 0.222 & 0.277 & 0.225 & 0.268 &0.236 &0.267 & 0.229 & 0.279 & 0.234 & 0.279 & 0.282 & 0.292 & 0.272 & 0.331 \\
\midrule
ETTh2      & \textcolor{red}{\textbf{0.338}} & \underline{\textcolor{second_blue}{0.383}} & 0.368 & 0.404 & 0.350 & 0.389 & 0.353 & 0.395 & 0.347 & 0.389 & 0.357 & 0.395 & \textcolor{red}{\textbf{0.338}} & \textcolor{red}{\textbf{0.377}} & 0.385 & 0.414 & 0.385 & 0.420 & 0.339 & 0.384 & 0.482 & 0.472 \\
\midrule
ETTm2     & \textcolor{red}{\textbf{0.257}} & \underline{\textcolor{second_blue}{0.315}} & 0.273 & 0.326 & 0.266 & 0.321 & 0.259 & 0.316 & 0.272 & 0.325 & 0.262 & 0.320 &\underline{\textcolor{second_blue}{0.258}} &\textcolor{red}{\textbf{0.309}} & 0.296 & 0.329 & 0.279 & 0.328 & 0.258 & 0.316 & 0.268 & 0.332 \\
\midrule
Exchange   & \underline{\textcolor{second_blue}{0.372}} & \underline{\textcolor{second_blue}{0.408}} & 0.423 & 0.443 & 0.373 & 0.414 & 0.403 & 0.422 & 0.404 & 0.422 & 0.385 & 0.419 &0.394 &0.415 & 0.416 & 0.442 & 0.432 & 0.435 & 0.382 & 0.424 & \textcolor{red}{\textbf{0.333}} & \textcolor{red}{\textbf{0.392}} \\

\bottomrule
\end{tabular}
}  
\caption{Multivariate time series forecasting results (averaged over all prediction lengths). For ILI and COVID-19, the look-back window is 48 and prediction lengths are $\{24, 36, 48, 60\}$. For all other datasets, the look-back window is 336 and prediction lengths are $\{96, 192, 336, 720\}$. Lower MSE/MAE indicates better performance. \textcolor{red}{\textbf{Bold red}} denotes the best result, \underline{\textcolor{blue}{blue underline}} denotes the second best. Full horizon-specific results are provided in Appendix~\ref{sec:appendix_full_results}.}
\label{tab:main_results}
\end{table}

\subsection{Mechanism Analysis}
\label{sec:mechanism_analysis}

\begin{figure}[t]
\centering
\includegraphics{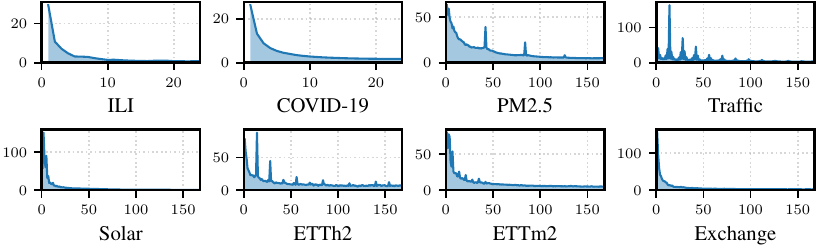}
\caption{Mean amplitude spectra of the eight datasets. Real-world time series exhibit intrinsically diverse spectral signatures, ranging from smooth non-periodic decay (e.g., COVID-19) and sparse high-amplitude concentration (e.g., Solar) to complex multi-frequency distributions (e.g., ETTh2), fundamentally driving RouteTS's routing decisions.
}
\label{fig:spectra_analysis}
\end{figure}

\begin{figure}[t]
\centering
\includegraphics{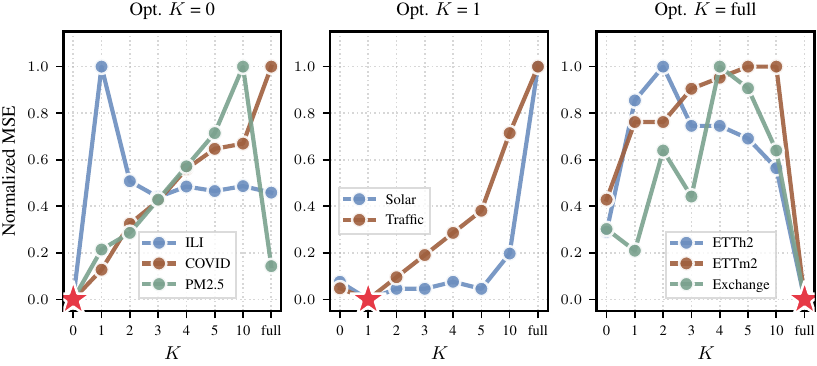}
\caption{Normalized MSE degradation across different routing hyperparameters ($K$). Red stars (\textcolor[HTML]{E63946}{$\star$}) indicate the global optimal $K$ for each dataset. RouteTS correctly anchors at these optima, while deviating from them incurs significant performance penalties due to misaligned inductive biases.}
\label{fig:k_sensitivity}
\end{figure}

\begin{figure}[t]
\centering
\includegraphics{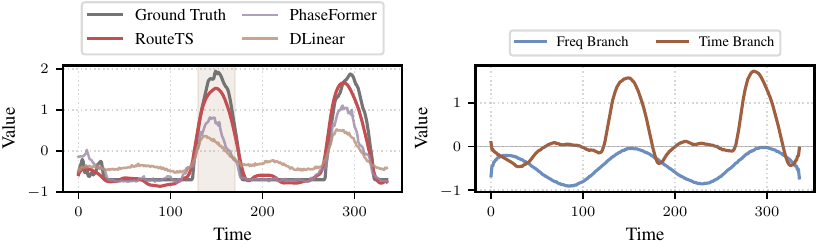}
\caption{Domain-routing behavior on Solar ($K=1$) under the predict-336 settings. Left: RouteTS accurately tracks periodic structures while time-domain baselines suffer from misalignment. Right: the Frequency Branch captures principal periodicities and the Time Branch models residual local variations.}
\label{fig:fig_pure_temporal_analysis_iclr}
\end{figure}

To understand how data characteristics drive RouteTS's routing decisions, we analyze the spectral properties of the datasets, as illustrated in Figure~\ref{fig:spectra_analysis}. The results demonstrate that RouteTS adapts its modeling strategy to match the data's intrinsic periodicity and spectral energy distribution. The routing behaviors fall into three distinct categories:

\begin{itemize}
    \item \textbf{Pure Time-Domain Dependency:} Datasets such as ILI and COVID-19 lack prominent periodic patterns. Their amplitude spectra decay smoothly without identifiable dominant frequencies, indicating the absence of stable cyclical structure that the Frequency Branch could exploit. Consequently, routing all information to the time domain yields the best performance. Furthermore, as shown in Figure~\ref{fig:k_sensitivity}, deviating from $K=0$ and allocating any spectral component to the frequency domain causes severe performance degradation. PM2.5 also selects $K=0$, yet the performance variation across all hyperparameter values is negligible (maximum MSE gap of 0.003). This insensitivity stems from the absence of predictable periodic patterns in both domains: the spectrum lacks concentrated energy, and the time series is dominated by stochastic fluctuations that neither branch can reliably capture. We provide a detailed Autocorrelation Function analysis of this phenomenon in Appendix~\ref{sec:appendix_acf}.
    
    \item \textbf{Single Dominant Frequency:} Traffic and Solar exhibit pronounced energy concentration in the frequency domain. Their spectra are dominated by one or a few high-amplitude frequencies that stand significantly above the spectral floor. Setting $K=1$ enables RouteTS to isolate this primary periodic component efficiently while delegating residual non-stationary variations to the Time Branch. Figure~\ref{fig:k_sensitivity} confirms that increasing the frequency allocation introduces high-frequency noise and degrades the prediction accuracy of the residual branch. Notably, deviating from $K=1$ toward either pure time-domain or pure frequency-domain operation incurs significant performance penalties, confirming that dual-branch collaboration is essential when the data exhibits a dominant yet non-exclusive periodic structure. Figure~\ref{fig:fig_pure_temporal_analysis_iclr} visualizes a representative Solar sample under the predict-336 settings. The left panel corroborates the periodic misalignment of time-domain baselines identified in Section~\ref{sec:intro}, while the right panel reveals the internal decomposition: the Frequency Branch captures the sparse high-amplitude periodicity, and the Time Branch compensates for local residual variations. The limitations of pure frequency-domain modeling on such mixed-regime data are further visualized in Appendix~\ref{sec:appendix_freq_oversmoothing}.
    
    \item \textbf{Complex Periodicity:} ETTh2 and ETTm2 contain broad spectral energy across multiple frequencies. Routing them entirely to the frequency domain by setting $K=\text{full}$ preserves macro-level periodic information and yields the best performance. Exchange exhibits a diffuse spectral energy distribution without prominent periodic spikes, yet empirically selects the full frequency-domain routing. Table~\ref{tab:time_operator_sensitivity} further indicates that the Time-MLP yields inferior performance on this dataset compared to the linear frequency-domain predictor. The constrained hypothesis space of the linear frequency operator effectively regularizes this highly volatile financial series, suppressing overfitting to local noise.
\end{itemize}

\begin{table}[htbp]
\centering
\caption{Operator sensitivity on Exchange. The linear frequency-domain predictor ($K=\text{full}$) outperforms the Time-MLP ($K=0$), indicating that the constrained hypothesis space of the linear operator provides effective regularization for highly volatile financial series.}
\label{tab:time_operator_sensitivity}
\resizebox{\textwidth}{!}{%
\begin{tabular}{@{}l c c l c@{}}
\toprule
Dataset & Time-MLP ($K=0$) & Time-Linear ($K=0$) & Change (MLP $\rightarrow$ Linear) & Opt. $K$ \\ \midrule
Exchange         & 0.398                     & 0.376                        & \textbf{Improve (-5.5\%)}                  & $\text{full}$ \\ \bottomrule
\end{tabular}%
}
\end{table}

\subsection{Model Efficiency Analysis}
\label{sec:efficiency_analysis}

\begin{figure}[t]
\centering
\includegraphics[width=\linewidth]{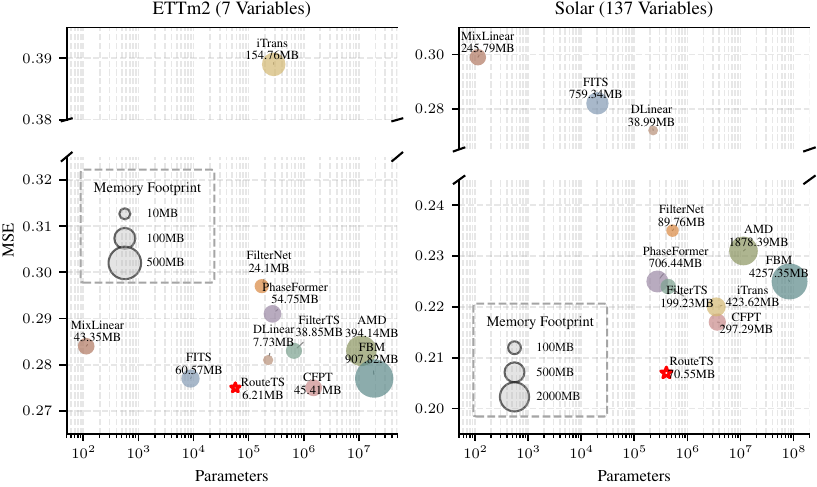}
\caption{Model effectiveness and efficiency comparison under extreme long-sequence settings (Look-back=336, Pred=336). Bubble size indicates maximum memory footprint. RouteTS delivers strong predictive performance with minimal memory and parameter overhead.}
\label{fig:efficiency_comparison}
\end{figure}

We stress-test computational efficiency under long-sequence conditions on ETTm2 and Solar. Figure~\ref{fig:efficiency_comparison} reports MSE, parameter count, and memory footprint.

Transformer-based models and dense frequency architectures suffer severe memory expansion due to $\mathcal{O}(L^2)$ attention complexity or exhaustive spectral operations. For example, AMD and FBM consume over 1.8GB and 4.2GB on Solar, respectively, limiting deployment on resource-constrained hardware. Meanwhile, minimalist linear methods such as MixLinear exhibit low memory usage but suffer from significant underfitting.

RouteTS breaks this capability--cost trade-off. On Solar, it achieves competitive MSE with merely 70.55MB of memory. On ETTm2, it matches CFPT's accuracy yet requires only 4\% of the parameters and 13\% of the memory. These results validate that amplitude routing inherently serves as a highly parameter-efficient sparse filter, enabling strong predictive performance without prohibitive overhead.

\section{CONCLUSION}

This paper identifies a fundamental limitation of single-domain forecasting: entangling global periodicities and non-stationary variations forces a trade-off between periodic misalignment and over-smoothing. To transcend this, we propose RouteTS, a framework that dynamically routes spectral energy based on amplitude. A complex-valued Frequency Branch preserves the amplitude-phase structure of dominant harmonics, while a lightweight MLP Time Branch captures non-linear residuals. Extensive evaluations reveal that optimal routing is strictly dictated by the data's intrinsic spectral signature. RouteTS achieves favorable accuracy across most benchmarks, with mechanism analyses confirming its resilience to long-range phase drift and mixed-regime instability. Crucially, these predictive gains are realized with minimal computational overhead. A current limitation is its channel-independent design; integrating inter-channel relational modeling represents a promising future direction.

\section*{AI Use Statement}

In this work, we used generative AI tools for drafting and editing portions of the research paper to improve readability, clarity, and flow, as well as for refining \LaTeX{} code for table and figure formatting.

We have not used generative AI tools for generating synthetic datasets, developing theoretical models or conceptual frameworks, formulating mathematical claims, providing critical ingredients for proofs, designing or implementing the proposed methodology or experiments, cleaning or reformatting datasets, supporting qualitative data analysis, or interpreting experimental results.

We have reviewed all AI-assisted work. All AI-generated text was carefully verified against the authors' original research contributions, experimental records, and manually implemented code. The core methodology, mathematical formulations, experimental design, and software implementation were developed independently by the authors without generative AI assistance. We take full responsibility for the final content of this work, including text, claims, or artifacts produced with the aid of generative AI.

\section*{Reproducibility Statement}

We provide comprehensive details to facilitate the reproduction of our results. The proposed RouteTS architecture and its data flow are fully specified in Section~\ref{subsec:amplitude_routing}--Section~\ref{subsec:time_branch}. Dataset statistics, preprocessing steps, and baseline descriptions are documented in Appendix~\ref{app:datasets} and Appendix~\ref{app:baselines}. All training hyperparameters, including dataset-specific batch sizes, learning rates, and routing thresholds $K$, are reported in Table~\ref{tab:hyperparams} and Appendix~\ref{app:implementation}. The main experimental protocol is detailed in Section~\ref{subsec:setup}. Complete per-horizon results are provided in Appendix~\ref{sec:appendix_full_results}.

\bibliography{iclr2027_conference}

@article{hu_adaptive_2025,
  title = {Adaptive {{Multi-Scale Decomposition Framework}} for {{Time Series Forecasting}}},
  author = {Hu, Yifan and Liu, Peiyuan and Zhu, Peng and Cheng, Dawei and Dai, Tao},
  year = 2025,
  month = apr,
  journal = {Proceedings of the AAAI Conference on Artificial Intelligence},
  volume = {39},
  number = {16},
  pages = {17359--17367},
  issn = {2374-3468, 2159-5399},
  doi = {10.1609/aaai.v39i16.33908}
}

@inproceedings{kou_cfpt_2025,
  title = {{{CFPT}}: {{Empowering}} Time Series Forecasting through Cross-Frequency Interaction and Periodic-Aware Timestamp Modeling},
  booktitle = {Proceedings of the 42nd International Conference on Machine Learning},
  author = {Kou, Feifei and Wang, Jiahao and Shi, Lei and others},
  year = 2025,
  pages = {31634--31647},
  publisher = {PMLR}
}

@inproceedings{liu2024itransformer,
  title = {{{iTransformer}}: {{Inverted}} Transformers Are Effective for Time Series Forecasting},
  booktitle = {International Conference on Learning Representations ({{ICLR}})},
  author = {Liu, Yong and Hu, Tengge and Zhang, Haoran and Wu, Haixu and Wang, Shiyu and Ma, Lintao and Long, Mingsheng},
  year = 2024
}

@article{wang_filterts_2025,
  title = {{{FilterTS}}: {{Comprehensive Frequency Filtering}} for {{Multivariate Time Series Forecasting}}},
  shorttitle = {{{FilterTS}}},
  author = {Wang, Yulong and Liu, Yushuo and Duan, Xiaoyi and Wang, Kai},
  year = 2025,
  month = apr,
  journal = {Proceedings of the AAAI Conference on Artificial Intelligence},
  volume = {39},
  number = {20},
  pages = {21375--21383},
  issn = {2374-3468, 2159-5399},
  doi = {10.1609/aaai.v39i20.35438}
}

@inproceedings{xu_fits_2024,
  title = {{{FITS}}: {{Modeling}} Time Series with {{10k}} Parameters},
  booktitle = {The Twelfth International Conference on Learning Representations},
  author = {Xu, Zhijian and Zeng, Ailing and Xu, Qiang},
  year = 2024
}

@inproceedings{yi_filternet_2024,
  title = {{{FilterNet}}: {{Harnessing}} Frequency Filters for Time Series Forecasting},
  booktitle = {Advances in Neural Information Processing Systems},
  author = {Yi, Kun and Fei, Jingru and Zhang, Qi and He, Hui and Hao, Shufeng and Lian, Defu and Fan, Wei},
  editor = {Globerson, A. and Mackey, L. and Belgrave, D. and Fan, A. and Paquet, U. and Tomczak, J. and Zhang, C.},
  year = 2024,
  volume = {37},
  pages = {55115--55140},
  publisher = {Curran Associates, Inc.},
  doi = {10.52202/079017-1749}
}

@article{zeng_are_2023,
  title = {Are {{Transformers Effective}} for {{Time Series Forecasting}}?},
  author = {Zeng, Ailing and Chen, Muxi and Zhang, Lei and Xu, Qiang},
  year = 2023,
  month = jun,
  journal = {Proceedings of the AAAI Conference on Artificial Intelligence},
  volume = {37},
  number = {9},
  pages = {11121--11128},
  issn = {2374-3468, 2159-5399},
  doi = {10.1609/aaai.v37i9.26317},
  langid = {english}
}

@inproceedings{runze2024rethinking,
  title = {Rethinking {{Fourier}} Transform from {{A}} Basis Functions Perspective for Long-Term Time Series Forecasting},
  booktitle = {The Thirty-Eighth Annual Conference on Neural Information Processing Systems},
  author = {Wu, Runze and Liang, Yuxuan and Song, Xuan and Wang, Kun and Lin, Youfang},
  year = 2024
}

@inproceedings{niu2026phaseformer,
  title = {{{PhaseFormer}}: {{From}} Patches to Phases for Efficient and Effective Time Series Forecasting},
  booktitle = {The Fourteenth International Conference on Learning Representations},
  author = {Niu, Yiming and Deng, Jinliang and Tong, Yongxin},
  year = 2026
}

@inproceedings{ma2026mixlinear,
  title = {{{MixLinear}}: {{Extreme}} Low Resource Multivariate Time Series Forecasting with \$0.{{1K}}\$ Parameters},
  booktitle = {The Fourteenth International Conference on Learning Representations},
  author = {Ma, Aitian and Luo, Dongsheng and Sha, Mo},
  year = 2026
}

@inproceedings{Adam,
  title = {Adam: {{A}} Method for Stochastic Optimization},
  booktitle = {3rd International Conference on Learning Representations, {{ICLR}} 2015, San Diego, {{CA}}, {{USA}}, May 7-9, 2015, Conference Track Proceedings},
  author = {Kingma, Diederik P. and Ba, Jimmy},
  editor = {Bengio, Yoshua and LeCun, Yann},
  year = 2015,
  bibsource = {dblp computer science bibliography, https://dblp.org}
}

@article{paszke_pytorch_,
  title = {{{PyTorch}}: {{An Imperative Style}}, {{High-Performance Deep Learning Library}}},
  author = {Paszke, Adam and Gross, Sam and Massa, Francisco and Lerer, Adam and Bradbury, James and Chanan, Gregory and Killeen, Trevor and Lin, Zeming and Gimelshein, Natalia and Antiga, Luca and Desmaison, Alban and Kopf, Andreas and Yang, Edward and DeVito, Zachary and Raison, Martin and Tejani, Alykhan and Chilamkurthy, Sasank and Steiner, Benoit and Fang, Lu and Bai, Junjie and Chintala, Soumith},
  langid = {english},
  year = 2019,
}

@inproceedings{nie2023time,
  title = {A Time Series Is Worth 64 Words: {{Long-term}} Forecasting with Transformers},
  booktitle = {International Conference on Learning Representations ({{ICLR}})},
  author = {Nie, Yuqi and Nguyen, Nam H. and Sinthupinyo, Phanwadee and Krishna, Satyapriya},
  year = 2023
}

@inproceedings{wu2021autoformer,
  title = {Autoformer: {{Decomposition}} Transformers with Auto-Correlation for Long-Term Series Forecasting},
  booktitle = {Advances in Neural Information Processing Systems},
  author = {Wu, Haixu and Xu, Jiehui and Wang, Jianmin and Long, Mingsheng},
  year = 2021,
  volume = {34},
  pages = {22413--22424}
}

@inproceedings{zhou2021informer,
  title = {Informer: {{Beyond}} Efficient Transformer for Long Sequence Time-Series Forecasting},
  booktitle = {Proceedings of the {{AAAI}} Conference on Artificial Intelligence},
  author = {Zhou, Haoyi and Zhang, Shanghang and Peng, Jieqi and Zhang, Shuai and Li, Jianxin and Xiong, Hui and Zhang, Wancai},
  year = 2021,
  volume = {35},
  pages = {11106--11115}
}

@article{chen2011energy,
  title = {Energy Time Series Forecasting Based on Pattern Sequence Similarity},
  author = {Chen, Z. F. and Aghakhani, S. and Man, J. and Dick, S.},
  year = 2011,
  journal = {IEEE Transactions on Knowledge and Data Engineering},
  volume = {23},
  number = {4},
  pages = {592--605},
  publisher = {IEEE}
}

@inproceedings{cirstea2021enhancenet,
  title = {{{EnhanceNet}}: {{Plugin}} Neural Networks for Enhancing Correlated Time Series Forecasting},
  booktitle = {2021 {{IEEE}} 37th International Conference on Data Engineering ({{ICDE}})},
  author = {Cirstea, Razvan-Gabriel and Kieu, Tung and Guo, Chenjuan and Yang, Bin and Pan, Sinno Jialin},
  year = 2021,
  pages = {1739--1750},
  publisher = {IEEE}
}

@article{ju2021novel,
  title = {A Novel Deep Class-Imbalanced Semisupervised Model for Wind Turbine Blade Icing Detection},
  author = {Ju, Meng and Zhao, Yue and Wang, Hao and Dong, Zhao and others},
  year = 2021,
  journal = {IEEE Transactions on Industrial Informatics},
  publisher = {IEEE}
}

@inproceedings{lai2018modeling,
  title = {Modeling Long- and Short-Term Temporal Patterns with Deep Neural Networks},
  booktitle = {The 41st International {{ACM SIGIR}} Conference on Research \& Development in Information Retrieval},
  author = {Lai, Guokun and Chang, Wei-Cheng and Yang, Yiming and Liu, Hanxiao},
  year = 2018,
  pages = {95--104}
}

@article{wu2022autocts,
  title = {{{AutoCTS}}: Automated Correlated Time Series Forecasting},
  author = {Wu, Xinle and Zhang, Dalin and Guo, Chenjuan and He, Chaoyang and Yang, Bin and Jensen, Christian S},
  year = 2022,
  journal = {Proceedings of the VLDB Endowment},
  volume = {15},
  number = {5},
  pages = {971--983},
  publisher = {VLDB Endowment}
}

@article{He2022InstanceBased,
  title     = {Instance-based deep transfer learning with attention for stock movement prediction},
  author    = {He, Qi-Qiao and Siu, Shirley Weng In and Si, Yain-Whar},
  journal   = {Knowledge and Information Systems},
  year      = {2022},
  volume    = {64},
  number    = {12},
  pages     = {3371--3395},
  doi       = {10.1007/s10489-022-03755-2},
  publisher = {Springer}
}

@book{madsen2007time-series-analysis,
  title={Time Series Analysis},
  author={Henrik Madsen},
  year={2007},
  publisher={CRC Press}
}

@article{ma_lstm_2015,
  title = {{{LSTM}} Network: A Deep Learning Approach for Short-Term Traffic Forecast},
  author = {Ma, Xiaolei and Tao, Zhimin and Wang, Yinhai and Yu, Hanjing and Wang, Yunpeng},
  year = 2015,
  journal = {IEEE Transactions on Intelligent Transportation Systems},
  volume = {16},
  number = {5},
  pages = {2985--2996},
  publisher = {IEEE}
}

@article{rangapuram_deep_2018,
  title = {Deep State Space Models for Time Series Forecasting},
  author = {Rangapuram, Syama Sundar and Seeger, Matthias W. and Gasthaus, Jan and Stella, Lorenzo and Wang, Yuyang and Januschowski, Tim},
  year = 2018,
  journal = {Advances in Neural Information Processing Systems},
  volume = {31}
}

@article{salinas_deepar_2020,
  title = {{{DeepAR}}: {{Probabilistic}} Forecasting with Autoregressive Recurrent Networks},
  author = {Salinas, David and Flunkert, Valentin and Gasthaus, Jan},
  year = 2020,
  journal = {International Journal of Forecasting},
  volume = {36},
  number = {3},
  pages = {1181--1191},
  publisher = {Elsevier}
}

@article{bai_empirical_2018,
  title = {An Empirical Evaluation of Generic Convolutional and Recurrent Networks for Sequence Modeling},
  author = {Bai, Shaojie and Kolter, J. Zico and Koltun, Vladlen},
  year = 2018,
  journal = {CoRR},
  volume = {abs/1803.01271},
  eprint = {1803.01271},
  archiveprefix = {arXiv},
  bibsource = {dblp computer science bibliography, https://dblp.org}
}

@article{liu_time_2021,
  title = {Time Series Is a Special Sequence: {{Forecasting}} with Sample Convolution and Interaction},
  author = {Liu, Minhao and Zeng, Ailing and Lai, Qiuxia and Xu, Qiang},
  year = 2021,
  journal = {CoRR},
  volume = {abs/2106.09305},
  eprint = {2106.09305},
  archiveprefix = {arXiv},
  bibsource = {dblp computer science bibliography, https://dblp.org}
}
\bibliographystyle{iclr2027_conference}

\appendix

\section{Experimental Details}

\subsection{Datasets}
\label{app:datasets}

We provides a comprehensive overview of the eight benchmark datasets utilized in our experiments. Table~\ref{tab:dataset_desc} summarizes the statistical properties of these datasets, including the number of input channels (variables), temporal sampling rates, and total available timesteps.

\begin{table}[h]
\centering
\caption{Statistical description of the eight benchmark datasets.}
\label{tab:dataset_desc}
\resizebox{\textwidth}{!}{%
\begin{tabular}{l|cccccccc}
\toprule
Dataset & ILI & COVID-19 & PM2.5 & Traffic & Solar & ETTh2 & ETTm2 & Exchange \\
\midrule
Channels & 7 & 8 & 184 & 862 & 137 & 7 & 7 & 8 \\
Sampling Rate & 1 week & 1 day & 3 hour & 1 hour & 10 min & 1 hour & 15 min & 1 day \\
Total Timesteps & 966 & 1,143 & 11,688 & 17,544 & 52,560 & 17,420 & 69,680 & 7,588 \\
\bottomrule
\end{tabular}%
}
\end{table}

\noindent\textbf{ILI Dataset\footnote{\url{https://github.com/juyongjiang/TimeSeriesDatasets}}.} 
The Influenza-like Illness (ILI) dataset comprises weekly reports on influenza-like cases monitored by the Centers for Disease Control and Prevention (CDC) in the United States. Spanning 966 timesteps with 7 clinical and demographic channels, it records the ratio of patients exhibiting influenza symptoms to the total patients treated. Modeling this dataset is crucial for public health decision-making, though its coarse weekly resolution and sparse timesteps make long-term forecasting exceptionally challenging.

\noindent\textbf{COVID-19 Dataset\footnote{\url{https://raw.githubusercontent.com/CSSEGISandData/COVID-19/master/csse_covid_19_data/csse_covid_19_time_series/time_series_covid19_confirmed_global.csv}}.} 
This dataset records daily cumulative confirmed cases across major global regions, sourced from the Johns Hopkins University (JHU) Center for Systems Science and Engineering. Rather than applying temporal differencing or smoothing, the dataset retains the raw cumulative time-series observations, preserving the original monotonic growth and non-stationary trends. It captures the top 8 countries with the highest cumulative cases at the end of the observation window, with the United States designated as the target variable ($OT$). Spanning 1,143 daily timesteps, this dataset evaluates the capacity of forecasting architectures to model continuous monotonic growth patterns and long-term distributional shifts.

\noindent\textbf{PM2.5 Dataset\footnote{\url{https://github.com/hqh0728/TimeBase}\label{fn:timebase}}.} 
The PM2.5 dataset monitors atmospheric fine particulate matter concentrations alongside regional meteorological indicators in Beijing, recorded at a 3-hour sampling interval. It consists of 11,688 timesteps across 184 spatial and environmental channels representing diverse monitoring stations. Characterized by its high dimensionality, intricate spatial-temporal correlations, and significant seasonal fluctuations under non-stationary dynamics, this dataset serves as a demanding benchmark for evaluating high-dimensional multivariate forecasting architectures.

\noindent\textbf{Traffic Dataset\textsuperscript{\ref{fn:timebase}}.} 
The Traffic dataset provides hourly road occupancy rates measured by 862 inductive loop sensors deployed along major freeways in the San Francisco Bay Area, maintained by the California Department of Transportation. Comprising 17,544 timesteps, the dataset exhibits strong daily and weekly periodicities intertwined with abrupt, non-stationary patterns caused by rush-hour congestion, making it a standard testbed for capturing spatial-temporal interactions.

\noindent\textbf{Solar Dataset\textsuperscript{\ref{fn:timebase}}.} 
The Solar-Energy dataset records solar power generation tracked at 10-minute intervals throughout 2016 across 137 photovoltaic (PV) plants in Alabama. It consists of 52,560 timesteps, capturing high-frequency fluctuations in solar irradiance alongside prominent diurnal and seasonal periodicities, which is ideal for validating the model's ability to capture global deterministic periodic components.

\noindent\textbf{Exchange Dataset\textsuperscript{\ref{fn:timebase}}.} 
The Exchange-Rate dataset tracks the daily currency exchange rates of eight major foreign economies (including Australia, Canada, Switzerland, China, Japan, New Zealand, Singapore, and the United Kingdom) relative to the US dollar from 1990 to 2016. Spanning 7,588 timesteps, this dataset is characterized by high non-stationarity, market volatility, and a lack of explicit temporal periodicities, serving as a demanding benchmark for economic financial series modeling.

\noindent\textbf{ETT Dataset\textsuperscript{\ref{fn:timebase}}.} 
The Electricity Transformer Temperature (ETT) datasets consist of key operational parameters and oil temperatures collected from electricity substations. In our experiments, we utilize two challenging subsets: ETTh2 (sampled hourly with 17,420 timesteps) and ETTm2 (sampled at 15-minute intervals with 69,680 timesteps). Both configurations comprise 7 variables representing transformer load and oil temperature indicators, providing a foundational benchmark for testing long-term forecasting across different temporal granularities.

\subsection{Baselines}
\label{app:baselines}
We select ten representative and state-of-the-art baselines for comparison, including Transformer-based models, Linear-based models, and Frequency-domain models. We introduce these baselines as follows:

\noindent\textbf{PhaseFormer.} PhaseFormer\citep{niu2026phaseformer} is a lightweight model that introduces a phase-based perspective to characterize time-series periodicity. By reframing series into compact phase tokens instead of patches, it maps features into a shared latent space and employs a low-rank routing mechanism for efficient cross-phase interactions. This design successfully alleviates the computational bottlenecks of patch-level processing with only around 1k parameters.

\noindent\textbf{MixLinear.} MixLinear\citep{ma2026mixlinear} is an ultra-lightweight forecasting framework that combines segment-based trend extraction in the time domain with adaptive complex-valued low-rank filtering in the frequency domain. It reduces parameter complexity from $\mathcal{O}(n^2)$ to $\mathcal{O}(n)$ by utilizing strategic segmentation and rank-constrained filters, achieving high forecasting accuracy with merely 0.1k parameters.

\noindent\textbf{CFPT.} CFPT\citep{kou_cfpt_2025} is a dual-branch framework that leverages frequency dynamics and timestamp patterns via a Cross-Frequency Interaction (CFI) branch and a Periodic-Aware Timestamp Modeling (PTM) branch. The CFI branch extracts synergistic relationships among different frequencies through feature fusion, while the PTM branch converts 1D timestamps into 2D tensors to capture intra- and inter-period correlations via 2D convolutions.

\noindent\textbf{FilterTS.} FilterTS\citep{wang_filterts_2025} is a frequency-domain model utilizing specialized filtering modules to extract complex periodic and trend patterns. It integrates a Static Global Filtering Module to capture stable periodic components and a Dynamic Cross-Variable Filtering Module to emphasize shared frequency features across variables. By executing operations directly in the frequency domain, it avoids costly time-domain convolutions, boosting efficiency and accuracy.

\noindent\textbf{AMD.}  AMD\citep{hu_adaptive_2025} is a MLP-based framework designed to resolve the multi-scale entanglement effect in time series. It decomposes sequences into temporal patterns across different scales via a Decomposable Mixing block, models both temporal and channel dependencies, and aggregates them adaptively. This approach successfully captures multi-scale temporal variations with superior efficiency.

\noindent\textbf{FBM.} FBM\citep{runze2024rethinking} is a plug-and-play mapping method that embeds the discrete Fourier transform with basis expansion to capture explicit time-frequency relationships. By representing real and imaginary frequency components as coefficients of cosine and sine basis functions, it addresses the inconsistent starting cycle and series length issues in existing Fourier-based methods, easily enhancing various DNN backbones.

\noindent\textbf{iTransformer.} iTransformer\citep{liu2024itransformer} is a Transformer-based architecture that applies attention and feed-forward networks on inverted dimensions. It embeds the time points of individual series as independent variate tokens, capturing multivariate correlations via self-attention and learning nonlinear representations with MLPs. This inverted design avoids computational explosion and enables the effective utilization of long lookback windows.

\noindent\textbf{FilterNet.} FilterNet\citep{yi_filternet_2024} is an efficient network that replaces traditional linear and attention mappings with learnable frequency filters. It incorporates plain and contextual shaping filters to extract key temporal patterns by selectively passing or attenuating specific frequency components. This design mitigates vulnerability to high-frequency noise while achieving full-spectrum utilization.

\noindent\textbf{FITS.} FITS\citep{xu_fits_2024} is an exceptionally lightweight frequency-domain model that operates by interpolating time series in the complex frequency domain. Using a complex-valued network, it simultaneously captures amplitude and phase information to yield a compact yet comprehensive representation of temporal data, delivering competitive forecasting accuracy with only about 10k parameters.

\noindent\textbf{DLinear.} DLinear\citep{zeng_are_2023} is a Linear-based model that incorporates a classic decomposition scheme to handle complex temporal dynamics. It splits the input sequence into trend-cyclical and seasonal components via moving average, applying individual linear layers to forecast each component. This simple design drastically reduces architectural complexity while maintaining competitive performance against Transformer-based models.

\subsection{Implementation Details}
\label{app:implementation}

The training process was capped at a maximum of 100 epochs, incorporating an early stopping mechanism with a patience of 12 epochs evaluated on the validation set to prevent overfitting. 

To accommodate the diverse data scales, channel dimensions, and convergence behaviors of the benchmarks, we employed dataset-specific batch sizes and learning rates. Specifically, the batch size was set to 64 for the PM2.5 dataset, and 32 for the majority of the benchmarks (ETTh2, ETTm2, ILI, and COVID-19). For high-dimensional datasets with numerous variates (Traffic and Solar), the batch size was reduced to 16, and further scaled down to 8 for the Exchange dataset. These hierarchical adjustments were necessary to ensure stable gradient updates while preventing potential out-of-memory issues across varying dimensionalities.

The initial learning rates were empirically tuned for optimal convergence on each specific dataset. We set the learning rate to $1 \times 10^{-2}$ for ILI; $5 \times 10^{-3}$ for ETTh1, ETTh2, and Traffic; $1 \times 10^{-3}$ for COVID-19 and Solar; and $5 \times 10^{-4}$ for Exchange and PM2.5.

\section{FULL RESULTS}
\label{sec:appendix_full_results}

\begin{table*}[t]
\centering
\scriptsize  
\setlength{\tabcolsep}{1.8pt}  
\renewcommand{\arraystretch}{1.1}  
\resizebox{\linewidth}{!}{
\begin{tabular}{>{\centering\arraybackslash}p{0.45cm} | c  *{11}{|cc}}
\toprule
\multicolumn{2}{c|}{\small Model}
& \multicolumn{2}{c}{\makecell{\textbf{RouteTS}\\(\textbf{our})}}
& \multicolumn{2}{c}{\makecell{PhaseFormer\\(2026)}}
& \multicolumn{2}{c}{\makecell{MixLinear\\(2026)}}
& \multicolumn{2}{c}{\makecell{CFPT\\(2025)}}
& \multicolumn{2}{c}{\makecell{FilterTS\\(2025)}}
& \multicolumn{2}{c}{\makecell{AMD\\(2025)}}
& \multicolumn{2}{c}{\makecell{FBM\\(2024)}}
& \multicolumn{2}{c}{\makecell{iTransformer\\(2024)}}
& \multicolumn{2}{c}{\makecell{FilterNet\\(2024)}}
& \multicolumn{2}{c}{\makecell{FITS\\(2024)}}
& \multicolumn{2}{c}{\makecell{DLinear\\(2023)}} \\
\cmidrule{1-24}
\cmidrule(lr){3-4} \cmidrule(lr){5-6} \cmidrule(lr){7-8} \cmidrule(lr){9-10} \cmidrule(lr){11-12}
\cmidrule(lr){13-14} \cmidrule(lr){15-16} \cmidrule(lr){17-18} \cmidrule(lr){19-20} \cmidrule(lr){21-22}
\cmidrule(lr){23-24}
\multicolumn{2}{c|}{\small Metric}
& MSE & MAE & MSE & MAE & MSE & MAE & MSE & MAE & MSE & MAE & MSE & MAE
& MSE & MAE & MSE & MAE & MSE & MAE & MSE & MAE & MSE & MAE \\
\midrule
\multirow{5}{*}[0.15em]{\rotatebox{90}{\small ILI$^{\phantom{F}}$}}
& 24  & \textcolor{red}{\textbf{1.658}} & \textcolor{red}{\textbf{0.787}} & 3.008 & 1.172 & 2.759 & 1.103 & 2.174 & 0.902 & 2.162 & 0.917 & 2.847 & 1.175 &2.310 &0.967 & 2.539 & 1.053 & 2.278 & 0.945 & 3.118 & 1.269 & 2.267 & 1.015 \\
& 36 & \textcolor{red}{\textbf{1.578}} & \textcolor{red}{\textbf{0.777}} & 2.618 & 1.075 & 2.505 & 1.052 & \underline{\textcolor{second_blue}{1.981}} & \underline{\textcolor{second_blue}{0.899}} & 2.100 & 0.926 & 2.897 & 1.211 & 2.208&0.952 & 2.508 & 1.055 & 2.190 & 0.921 & 3.164 & 1.274 & 2.339 & 1.000 \\
& 48 & \textcolor{red}{\textbf{1.557}} & \textcolor{red}{\textbf{0.796}} & 2.690 & 1.100 & 2.352 & 1.020 & 2.171 & 0.953 & 2.145 & 0.951 & 3.243 & 1.301 & 2.167&0.948 & 2.571 & 1.077 & 2.230 & 0.928 & 2.674 & 1.110 & 2.456 & 1.050 \\
& 60 & \textcolor{red}{\textbf{1.770}} & \textcolor{red}{\textbf{0.867}} & 2.588 & 1.107 & 2.447 & 1.057 & 2.252 & 0.976 & 2.121 & 0.951 & 3.351 & 1.355 &2.136 &0.957 & 2.699 & 1.116 & 2.230 & 0.937 & 2.644 & 1.108 & 2.645 & 1.121 \\
\cmidrule(l){2-24}
& Avg & \textcolor{red}{\textbf{1.640}} & \textcolor{red}{\textbf{0.807}} & 2.726 & 1.114 & 2.516 & 1.058 & 2.144 & \underline{\textcolor{second_blue}{0.933}} & 2.132 & 0.936 & 3.085 & 1.261 & \underline{\textcolor{second_blue}{2.205}} &0.956 & 2.579 & 1.075 & 2.232 & \underline{\textcolor{second_blue}{0.933}} & 2.900 & 1.190 & 2.427 & 1.047 \\
\midrule
\multirow{5}{*}[0.15em]{\rotatebox{90}{\small COVID-19$^{\phantom{F}}$}}
& 24  & \textcolor{red}{\textbf{0.021}} & \textcolor{red}{\textbf{0.070}} & 0.062 & 0.107 & 0.308 & 0.251 & \underline{\textcolor{second_blue}{0.031}} & \underline{\textcolor{second_blue}{0.075}} & 0.251 & 0.211 & 0.678 & 0.406 &0.185 &0.150 & 0.294 & 0.234 & 0.056 & 0.113 & 0.226 & 0.202 & 0.062 & 0.163 \\
& 36 & \textcolor{red}{\textbf{0.053}} & \textcolor{red}{\textbf{0.096}} & 0.175 & 0.185 & 0.566 & 0.313 & \underline{\textcolor{second_blue}{0.067}} & \underline{\textcolor{second_blue}{0.110}} & 0.257 & 0.238 & 1.058 & 0.513 &0.515 &0.260 & 0.697 & 0.375 & 0.138 & 0.175 & 0.500 & 0.310 & 0.112 & 0.225 \\
& 48 & \textcolor{red}{\textbf{0.146}} & \textcolor{red}{\textbf{0.158}} & 0.216 & 0.202 & 0.781 & 0.385 & \underline{\textcolor{second_blue}{0.169}} & \underline{\textcolor{second_blue}{0.172}} & 0.303 & 0.271 & 1.567 & 0.632 &1.242 &0.412 & 1.147 & 0.499 & 0.314 & 0.289 & 0.889 & 0.439 & 0.178 & 0.300 \\
& 60 & \textcolor{red}{\textbf{0.280}} & \textcolor{red}{\textbf{0.217}} & 0.390 & \underline{\textcolor{second_blue}{0.273}} & 3.096 & 0.933 & 1.396 & 0.573 & 0.431 & 0.318 & 2.042 & 0.721 &2.935 &0.630 & 1.632 & 0.615 & 0.577 & 0.357 & 1.443 & 0.585 & \underline{\textcolor{second_blue}{0.376}} & 0.443 \\
\cmidrule(l){2-24}
& Avg & \textcolor{red}{\textbf{0.125}} & \textcolor{red}{\textbf{0.135}} & 0.211 & \underline{\textcolor{second_blue}{0.192}} & 1.188 & 0.470 & 0.416 & 0.233 & 0.310 & 0.260 & 1.336 & 0.568 &1.219 &0.363 & 0.943 & 0.431 & 0.271 & 0.234 & 0.764 & 0.384 & \underline{\textcolor{second_blue}{0.182}} & 0.283 \\
\midrule
\multirow{5}{*}[0.15em]{\rotatebox{90}{\small PM2.5$^{\phantom{F}}$}}
& 96 & 0.431 & \underline{\textcolor{second_blue}{0.428}} & 0.432 & 0.431 & 0.431 & 0.437 & 0.428 & 0.431 & 0.435 & 0.434 & 0.441 & 0.435 &0.449 &0.433 & 0.432 & 0.434 & 0.459 & 0.451 & \textcolor{red}{\textbf{0.421}} & \textcolor{red}{\textbf{0.424}} & \underline{\textcolor{second_blue}{0.422}} & 0.448 \\
& 192 & \underline{\textcolor{second_blue}{0.422}} & 0.422 & 0.437 & 0.432 & 0.430 & 0.436 & 0.427 & \textcolor{red}{\textbf{0.429}} & 0.433 & 0.432 & 0.436 & 0.431 &0.449 &0.431 & 0.430 & \textcolor{red}{\textbf{0.429}}& 0.446 & 0.442 & 0.426 & \underline{\textcolor{second_blue}{0.430}} & \textcolor{red}{\textbf{0.418}} & 0.445 \\
& 336 & \underline{\textcolor{second_blue}{0.412}} & \textcolor{red}{\textbf{0.420}} & 0.430 & 0.430 & 0.424 & 0.434 & 0.416 & \underline{\textcolor{second_blue}{0.424}} & 0.436 & 0.430 & 0.444 & 0.431 & 0.444&0.431 & 0.417 & 0.426& 0.427 & 0.428 & 0.422 & 0.430 & \textcolor{red}{\textbf{0.401}} & 0.448 \\
& 720 & \underline{\textcolor{second_blue}{0.401}} & 0.425 & 0.409 & 0.424 & 0.420 & 0.447 & 0.406 & 0.426 & \textcolor{red}{\textbf{0.379}} & \textcolor{red}{\textbf{0.413}} & 0.424 & 0.428 &0.418 &0.427 & \underline{\textcolor{second_blue}{0.401}} & \underline{\textcolor{second_blue}{0.419}} & 0.415 & 0.439 & 0.419 & 0.446 & 0.440 & 0.525 \\
\cmidrule(l){2-24}
& Avg & \textcolor{red}{\textbf{0.416}} & \textcolor{red}{\textbf{0.424}} & 0.427 & 0.429 & 0.426 & 0.438 & 0.419 & 0.428 & 0.421 & 0.427 & 0.436 & 0.431 &0.440 &0.430 & \underline{\textcolor{second_blue}{0.420}} & \underline{\textcolor{second_blue}{0.427}} & 0.437 & 0.440 & 0.422 & 0.432 & 0.420 & 0.467 \\
\midrule
\multirow{5}{*}[0.15em]{\rotatebox{90}{\small Traffic$^{\phantom{F}}$}}
& 96  & 0.373 & \underline{\textcolor{second_blue}{0.259}} & \underline{\textcolor{second_blue}{0.366}} & \textcolor{red}{\textbf{0.242}} & 0.416 & 0.281 & 0.376 & 0.257 & 0.382 & 0.272 & 0.392 & 0.275 &0.378 &0.258 & \textcolor{red}{\textbf{0.359}} & 0.260 & 0.370 & 0.267 & 0.411 & 0.281 & 0.429 & 0.314 \\
& 192 & 0.394 & 0.269 & \underline{\textcolor{second_blue}{0.387}} & \textcolor{red}{\textbf{0.252}} & 0.430 & 0.286 & 0.403 & \underline{\textcolor{second_blue}{0.266}} & 0.400 & 0.279 & 0.411 & 0.285 & 0.401&0.274 & \textcolor{red}{\textbf{0.375}} & 0.269 & 0.394 & 0.278 & 0.424 & 0.285 & 0.446 & 0.324 \\
& 336 & 0.406 & \underline{\textcolor{second_blue}{0.274}} & \underline{\textcolor{second_blue}{0.398}} & \textcolor{red}{\textbf{0.256}} & 0.444 & 0.296 & 0.416 & 0.279 & 0.414 & 0.286 & 0.425 & 0.290 &0.415 &0.280 & \textcolor{red}{\textbf{0.389}} & 0.275 & 0.407 & 0.283 & 0.436 & 0.291 & 0.458 & 0.329 \\
& 720 & \underline{\textcolor{second_blue}{0.436}} & 0.292 & 0.438 & \textcolor{red}{\textbf{0.282}} & 0.474 & 0.314 & 0.446 & 0.295 & 0.449 & 0.306 & 0.450 & 0.305 &0.452 &0.296 & \textcolor{red}{\textbf{0.421}} & \underline{\textcolor{second_blue}{0.291}} & 0.439 & 0.300 & 0.464 & 0.307 & 0.484 & 0.343 \\
\cmidrule(l){2-24}
& Avg & 0.402 & \underline{\textcolor{second_blue}{0.274}} & \underline{\textcolor{second_blue}{0.397}} & \textcolor{red}{\textbf{0.258}} & 0.441 & 0.294 & 0.410 & 0.274 & 0.411 & 0.286 & 0.420 & 0.289 & 0.412& 0.277& \textcolor{red}{\textbf{0.386}} & \underline{\textcolor{second_blue}{0.274}} & 0.403 & 0.282 & 0.434 & 0.291 & 0.454 & 0.328 \\
\midrule
\multirow{5}{*}[0.15em]{\rotatebox{90}{\small Solar$^{\phantom{F}}$}}
& 96  & \underline{\textcolor{second_blue}{0.186}} & \underline{\textcolor{second_blue}{0.236}} & 0.192 & \textcolor{red}{\textbf{0.231}} & 0.253 & 0.275 & 0.190 & 0.234 & \textcolor{red}{\textbf{0.183}} & 0.240 & 0.206 & 0.241 &0.202 &0.238 & 0.202 & 0.254 & 0.201 & 0.246 & 0.228 & 0.260 & 0.220 & 0.287 \\
& 192 & \textcolor{red}{\textbf{0.201}} &\underline{\textcolor{second_blue}{0.248}}& \underline{\textcolor{second_blue}{0.212}} & \textcolor{red}{\textbf{0.247}} & 0.282 & 0.308 & 0.210 & 0.254 & 0.217 & 0.268 & 0.225 & 0.254& 0.224& 0.265& 0.219 & 0.266 & 0.225 & 0.272 & 0.260 & 0.278 & 0.254 & 0.315 \\
& 336 & \textcolor{red}{\textbf{0.207}} & \underline{\textcolor{second_blue}{0.254}} & 0.225 & \textcolor{red}{\textbf{0.251}} & 0.299 & 0.307 & \underline{\textcolor{second_blue}{0.217}}& 0.263 & 0.224 & 0.276 & 0.231 & 0.263 &0.225 &0.259 & 0.220 & 0.268 & 0.235 & 0.279 & 0.282 & 0.292 & 0.272 & 0.329 \\
& 720 & \textcolor{red}{\textbf{0.211}} & \underline{\textcolor{second_blue}{0.258}} & \underline{\textcolor{second_blue}{0.217}} & \textcolor{red}{\textbf{0.252}} & 0.290 & 0.298 & 0.221 & 0.270 & 0.222 & 0.277 & 0.225 & 0.268 &0.236 &0.267 & 0.229 & 0.279 & 0.234 & 0.279 & 0.282 & 0.292 & 0.272 & 0.331 \\
\cmidrule(l){2-24}
& Avg & \textcolor{red}{\textbf{0.201}} & \underline{\textcolor{second_blue}{0.249}} & \underline{\textcolor{second_blue}{0.211}} & \textcolor{red}{\textbf{0.245}} & 0.281 & 0.297 & 0.210 & 0.255 & 0.212 & 0.265 & 0.222 & 0.257 &0.222 &0.257 & 0.217 & 0.267 & 0.224 & 0.269 & 0.263 & 0.281 & 0.254 & 0.316 \\
\midrule
\multirow{5}{*}[0.15em]{\rotatebox{90}{\small ETTh2$^{\phantom{F}}$}}
& 96  & \underline{\textcolor{second_blue}{0.275}} & \underline{\textcolor{second_blue}{0.337}} & 0.304 & 0.361 & 0.292 & 0.348 & \textcolor{red}{\textbf{0.274}} & \underline{\textcolor{second_blue}{0.337}} & \textcolor{red}{\textbf{0.274}} & \underline{\textcolor{second_blue}{0.337}} & 0.282 & 0.342 & 0.278 & \textcolor{red}{\textbf{0.331}} & 0.304 & 0.360 & 0.311 & 0.365 & \underline{\textcolor{second_blue}{0.275}} & \underline{\textcolor{second_blue}{0.337}} & 0.300 & 0.365 \\
& 192 & \underline{\textcolor{second_blue}{0.336}} & \textcolor{red}{\textbf{0.377}} & 0.371 & 0.398 & 0.350 & 0.380 & 0.342 & 0.384 & 0.342 & \underline{\textcolor{second_blue}{0.381}} & 0.360 & 0.389 & 0.349 & \textcolor{red}{\textbf{0.377}} & 0.392 & 0.413 & 0.375 & 0.413 & \textcolor{red}{\textbf{0.335}} & \textcolor{red}{\textbf{0.377}} & 0.407 & 0.434 \\
& 336 &\underline{\textcolor{second_blue}{0.355}} & \underline{\textcolor{second_blue}{0.396}} & 0.383 & 0.413 & 0.369 & 0.401 & 0.369 & 0.409 & 0.369 & 0.404 & 0.372 & 0.407 &\textcolor{red}{\textbf{0.344}} &\textcolor{red}{\textbf{0.386}} & 0.426 & 0.438 & 0.392 & 0.430 & 0.360 & 0.398 & 0.469 & 0.478 \\
& 720 & \underline{\textcolor{second_blue}{0.384}} & \underline{\textcolor{second_blue}{0.422}} & 0.415 & 0.443 & 0.392 & 0.427 & 0.427 & 0.450 & 0.403 & 0.435 & 0.416 & 0.443 & \textcolor{red}{\textbf{0.382}}& \textcolor{red}{\textbf{0.416}}& 0.416 & 0.443 & 0.460 & 0.473 & 0.386 & 0.424 & 0.752 & 0.609 \\
\cmidrule(l){2-24}
& Avg & \textcolor{red}{\textbf{0.338}} & \underline{\textcolor{second_blue}{0.383}} & 0.368 & 0.404 & 0.350 & 0.389 & 0.353 & 0.395 & 0.347 & 0.389 & 0.357 & 0.395 & \textcolor{red}{\textbf{0.338}}&\textcolor{red}{\textbf{0.377}} & 0.385 & 0.414 & 0.385 & 0.420 &0.339& 0.384 & 0.482 & 0.472 \\
\midrule
\multirow{5}{*}[0.15em]{\rotatebox{90}{\small ETTm2$^{\phantom{F}}$}}
& 96  & \textcolor{red}{\textbf{0.165}} & 0.255 & 0.173 & 0.261 & 0.178 & 0.267 & \underline{\textcolor{second_blue}{0.167}} & \underline{\textcolor{second_blue}{0.254}} & 0.176 & 0.260 & 0.173 & 0.261 &\textcolor{red}{\textbf{0.165}} &\textcolor{red}{\textbf{0.247}} & 0.180 & 0.272 & 0.191 & 0.270 & 0.167 & 0.256 & 0.168 & 0.262 \\
& 192 & \textcolor{red}{\textbf{0.221}} & \underline{\textcolor{second_blue}{0.292}} & 0.249 & 0.308 & 0.230 & 0.298 & 0.228 & 0.295 & 0.232 & 0.301 & 0.232 & 0.301 & \underline{\textcolor{second_blue}{0.223}} &\textcolor{red}{\textbf{0.286}} & 0.239 & 0.311 & 0.250 & 0.307 & 0.222 & 0.295 & 0.224 & 0.303 \\
& 336 &  \textcolor{red}{\textbf{0.275}} & 0.328 & 0.291 & 0.340 & 0.284 & 0.332 & \textcolor{red}{\textbf{0.275}} & \underline{\textcolor{second_blue}{0.327}} & 0.283 & 0.335 & 0.283 & 0.333 &\underline{\textcolor{second_blue}{0.277}}& \textcolor{red}{\textbf{0.323}} & 0.389 & 0.341 & 0.297 & 0.338 & 0.277 & 0.329 & 0.281 & 0.342 \\
& 720 &  0.368 & 0.383 & 0.380 & 0.394 & 0.373 & 0.386 & \underline{\textcolor{second_blue}{0.366}} & 0.386 & 0.396 & 0.402 & \textcolor{red}{\textbf{0.361}} & 0.384 &0.368 &\textcolor{red}{\textbf{0.379}} & 0.374 & 0.392 & 0.378 & 0.397 & 0.366 & \underline{\textcolor{second_blue}{0.382}} & 0.397 & 0.421 \\
\cmidrule(l){2-24}
& Avg & \textcolor{red}{\textbf{0.257}} & \underline{\textcolor{second_blue}{0.315}} & 0.273 & 0.326 & 0.266 & 0.321 & 0.259 & 0.316 & 0.272 & 0.325 & 0.262 & 0.320 &\underline{\textcolor{second_blue}{0.258}} &\textcolor{red}{\textbf{0.309}} & 0.296 & 0.329 & 0.279 & 0.328 & 0.258 & 0.316 & 0.268 & 0.332 \\
\midrule
\multirow{5}{*}[0.15em]{\rotatebox{90}{\small Exchange$^{\phantom{F}}$}}
& 96  & \textcolor{red}{\textbf{0.084}} & \textcolor{red}{\textbf{0.201}} & 0.102 & 0.227 & 0.101 & 0.227 & 0.090 & 0.212 & 0.095 & 0.217 & 0.098 & 0.224 & 0.093&0.210 & 0.099 & 0.226 & 0.100 & 0.225 & 0.096 & 0.214 & \underline{\textcolor{second_blue}{0.089}} & \underline{\textcolor{second_blue}{0.208}} \\
& 192 & \underline{\textcolor{second_blue}{0.175}} & \underline{\textcolor{second_blue}{0.297}} & 0.220 & 0.337 & 0.190 & 0.313 & 0.191 & 0.312 & 0.192 & 0.312 & 0.193 & 0.315 &0.192 &0.307 & 0.216 & 0.337 & 0.210 & 0.327 & 0.196 & 0.309 & \textcolor{red}{\textbf{0.156}} & \textcolor{red}{\textbf{0.288}} \\
& 336 & \underline{\textcolor{second_blue}{0.321}} & \underline{\textcolor{second_blue}{0.411}} & 0.397 & 0.462 & 0.329 & 0.417 & 0.360 & 0.434 & 0.387 & 0.446 & 0.346 & 0.426 & 0.345&0.419 & 0.387 & 0.461 & 0.367 & 0.436 & 0.380 & 0.420 & \textcolor{red}{\textbf{0.272}} & \textcolor{red}{\textbf{0.394}} \\
& 720 & 0.910 & \underline{\textcolor{second_blue}{0.725}} & 0.972 & 0.744 & 0.871 & 0.700 & 0.971 & 0.732 & 0.943 & 0.714 & 0.902 & 0.711 &0.946 &0.726 & 0.962 & 0.745 & 1.050 & 0.754 & \underline{\textcolor{second_blue}{0.854}} & 0.752 & \textcolor{red}{\textbf{0.814}} & \textcolor{red}{\textbf{0.678}} \\
\cmidrule(l){2-24}
& Avg & \underline{\textcolor{second_blue}{0.372}} & \underline{\textcolor{second_blue}{0.408}} & 0.423 & 0.443 & 0.373 & 0.414 & 0.403 & 0.422 & 0.404 & 0.422 & 0.385 & 0.419 &0.394 &0.415 & 0.416 & 0.442 & 0.432 & 0.435 & 0.382 & 0.424 & \textcolor{red}{\textbf{0.333}} & \textcolor{red}{\textbf{0.392}} \\

\midrule
\multicolumn{2}{c|}{\textbf{1\textsuperscript{st} Count}} & \textcolor{red}{\textbf{22}} & \textcolor{red}{\textbf{14}} & 0 & 10 & 0 & 0 & 2 & 1 & 3 & 1 & 1 & 0 & 5 & \underline{\textcolor{second_blue}{11}} & 5 & 1 & 0 & 0 & 2 & 1 & \underline{\textcolor{second_blue}{6}} & 4 \\

\bottomrule
\end{tabular}
}
\caption{Full multivariate time series forecasting results for all prediction lengths. Avg means the average of all prediction lengths.}
\label{tab:appendix_full_main_results}
\end{table*}

Table~\ref{tab:appendix_full_main_results} provides the complete multivariate forecasting results across all prediction lengths for each dataset. These results complement the averaged metrics reported in Table~\ref{tab:main_results}, offering a granular view of model performance under varying forecast horizons.

\section{Mechanism Analysis Supplement}

\subsection{Autocorrelation Function Analysis}

\label{sec:appendix_acf}

To corroborate the spectral analysis presented in the main text, we compute the Autocorrelation Function (ACF)\citep{madsen2007time-series-analysis} for the training sets of all evaluated benchmark datasets. The ACF provides a mathematical measure of cyclical predictability and memory retention in the time domain by quantifying the linear dependence between a time series and its lagged values. For a given time series $x$ of length $N$ with a mean of $\bar{x}$, the autocorrelation at lag $k$ is defined as:
\begin{equation}
    ACF(k) = \frac{\sum_{t=1}^{N-k} (x_t - \bar{x})(x_{t+k} - \bar{x})}{\sum_{t=1}^N (x_t - \bar{x})^2}
\end{equation}

\begin{figure}[htbp]
\centering
\includegraphics[width=\textwidth]{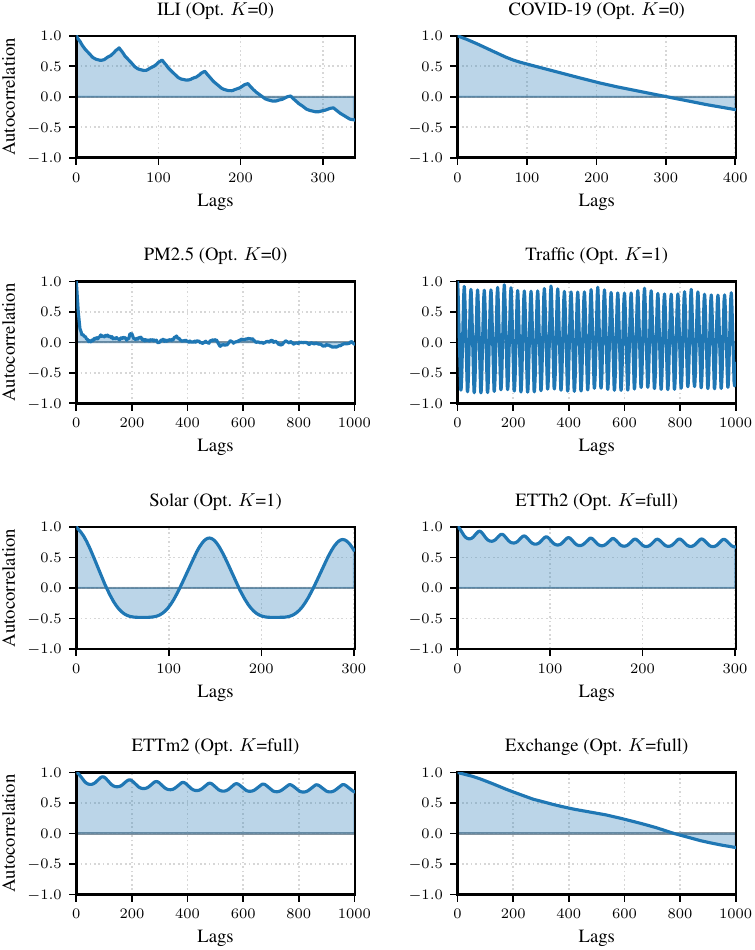}
\caption{Autocorrelation Function analysis computed on the training sets. Shaded areas represent the signal envelope. The visualization further corroborates the spectral analysis by revealing distinct periodic oscillations, smooth temporal decay, or rapid noise-bound drops across different datasets.}
\label{fig:acf_appendix}
\end{figure}

We employ the ACF as a diagnostic tool to validate the intrinsic temporal dynamics driving our routing mechanism. Specifically, the decay profile and the prominence of cyclical peaks in the ACF inherently dictate the optimal routing strategy. As illustrated in Figure~\ref{fig:acf_appendix}, the visual progression of the datasets aligns perfectly with the three routing archetypes identified in the main text:

\begin{itemize}
    \item \textbf{Pure Time-Domain Dependency:} Datasets in the first sequence, such as ILI and COVID-19, demonstrate a gradual, monotonic decline in autocorrelation without any cyclical peaks. This profile indicates the presence of long-term memory driven by continuous non-linear trends. PM2.5, while also selecting $K=0$, exhibits a rapid noise-bound drop where the autocorrelation collapses to near-zero immediately after the initial lags. Both profiles lack stable deterministic periodicity, entirely justifying RouteTS's decision to bypass the frequency domain.
    \item \textbf{Single Dominant Frequency:} Traffic and Solar exhibit perfectly undamped, high-amplitude sinusoidal oscillations across extended lags. The lack of significant baseline attenuation in their ACF curves confirms the existence of a highly deterministic, single-period cyclical pattern. This extreme correlation strictly supports the routing decision to isolate the primary harmonic by setting $K=1$, preventing high-frequency noise from interfering with the time domain.
    \item \textbf{Complex Periodicity and Regularization:} ETTh2 and ETTm2 present complex rippling patterns in their ACF, where high-frequency cyclical peaks are superimposed on a slower decay envelope. This indicates multi-scale periodicity, requiring the full frequency spectrum to capture macro-level temporal patterns. Conversely, Exchange shows a smooth decay resembling a random walk. As established in our main analysis, its allocation to $K=\text{full}$ is driven by the necessity for structural regularization against the Time-MLP operator, rather than cyclical extraction.
    
\end{itemize}

\subsection{Visualization of Frequency-Domain Over-Smoothing}
\label{sec:appendix_freq_oversmoothing}

While Figure~\ref{fig:fig_pure_temporal_analysis_iclr} demonstrates that time-domain models suffer from periodic misalignment on periodic-dominant data, we further visualize why pure frequency-domain models fail when localized transient variations coexist with dominant periodicities.

\begin{figure}[t]
\centering
\includegraphics{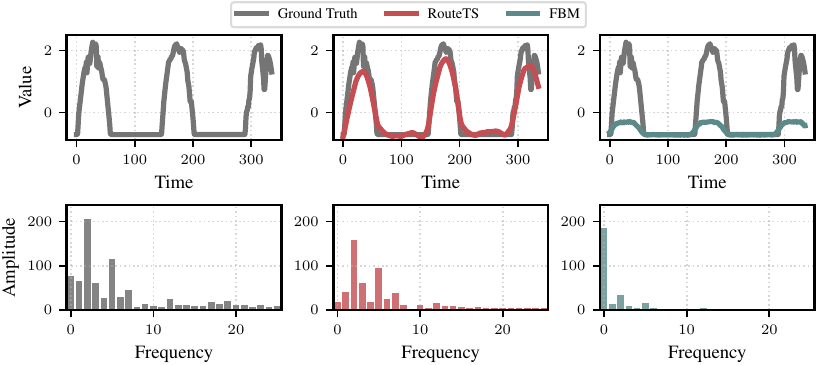}
\caption{Pure frequency-domain models suffer from over-smoothing and spectral energy collapse on mixed-regime data under the predict-336 settings. Top row: time-domain reconstructions of a representative Solar sample. Bottom row: corresponding amplitude spectra.}
\label{fig:fig_pure_freq_analysis_iclr}
\end{figure}

\textbf{Time-domain over-smoothing.} As shown in the top row of Figure~\ref{fig:fig_pure_freq_analysis_iclr}, FBM severely over-smooths the reconstructed waveform, collapsing the sharp peaks and valleys present in the ground truth into flattened, low-amplitude oscillations. In contrast, RouteTS accurately tracks the ground-truth trajectory by preserving both the principal periodic structure and the fine-grained local fluctuations.

\textbf{Spectral cause: excessive energy concentration.} The bottom row reveals the spectral origin of this over-smoothing. While the ground truth exhibits a broad distribution of energy across numerous frequency components---including both dominant low-frequency modes and non-negligible mid-to-high-frequency contributions---FBM's spectral energy collapses into an extremely sparse subset of low-frequency components. This excessive concentration indicates that FBM's frequency-domain representation fails to encode the rich spectral diversity of the original signal. Consequently, the inverse reconstruction lacks the high-frequency content necessary to recover sharp temporal transitions, inevitably producing the over-smoothed time-domain output observed above. RouteTS, by contrast, maintains a spectral profile closer to the ground truth: the Frequency Branch captures the dominant periodic modes, while the Time Branch reintroduces the residual broadband energy corresponding to localized variations, jointly preserving the full spectral-temporal structure.

\section{Ablation Study}

\subsection{Predictor Design in Time Branch}

To validate the architectural choice of the time-domain predictor, we compare a simple Linear mapping against the MLP backbone detailed in Section~\ref{subsec:time_branch} across four representative datasets spanning distinct routing configurations Traffic and Solar ($K=1$) and PM2.5 and COVID-19 ($K=0$).

\begin{figure}[htbp]
\centering
\includegraphics{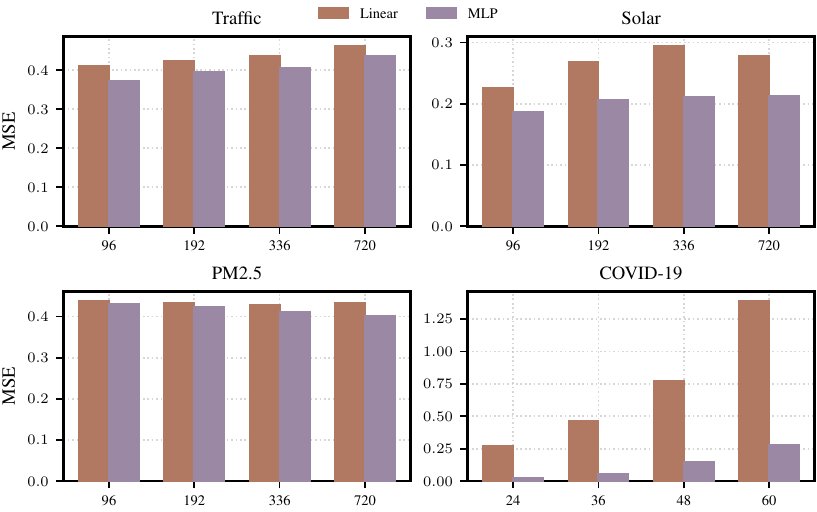}
\caption{Ablation study on the architectural design of the time-domain residual predictor.}
\label{fig:ablation_predictor}
\end{figure}

Figure~\ref{fig:ablation_predictor} reveals a consistent trend: the MLP outperforms the Linear baseline on all evaluated datasets and prediction horizons. The margin, however, is strictly governed by the data's non-stationarity and residual complexity.

\textbf{Highly non-stationary regimes.} On COVID-19, where the series exhibits continuous monotonic growth and abrupt distributional shifts, the MLP yields drastic improvements. As the horizon extends, the Linear predictor's MSE escalates from 0.28 to 1.35, whereas the MLP remains stable below 0.30. This confirms that non-linear activation is essential for modeling chaotic, high-frequency residuals that defy linear extrapolation.

\textbf{Hybrid routing with residual non-stationarity.} On Traffic and Solar ($K=1$), the Frequency Branch captures the dominant periodic structure, yet the Time Branch must still handle localized non-stationary deviations. The MLP consistently achieves lower MSE than the Linear variant across all horizons, indicating that even frequency-isolated residuals retain non-linear temporal dependencies that benefit from increased model capacity.

\textbf{Low signal-to-noise regimes.} On PM2.5 ($K=0$), where the spectrum lacks prominent periodicity and the signal is noise-dominated, the performance gap between MLP and Linear is narrow (within 0.01 MSE). Nevertheless, the MLP maintains a consistent edge across all horizons without overfitting, suggesting that its non-linear inductive bias extracts weak local structures without incurring measurable degradation.

These results justify the adoption of the MLP as the default time-domain predictor in RouteTS. By coupling it with dataset-specific routing ($K=0$ or $K=1$), the framework achieves optimal capacity allocation: the MLP captures complex local dynamics when needed, while the routing mechanism prevents unnecessary frequency-domain computation.

\subsection{Frequency Predictor Design}
\label{sec:freq_ablation}

We investigate the architectural necessity of the complex-valued linear mapping in the Frequency Branch. Our default design applies a shared learnable complex weight matrix $\mathbf{W} = \mathbf{W}_R + j\mathbf{W}_I$ to the historical spectrum. Following the algebraic rules of complex multiplication, the future spectrum is computed as
\[
\hat{\mathbf{X}}_{\text{freq}} = (\mathbf{W}_R + j\mathbf{W}_I)(\mathbf{X}_R + j\mathbf{X}_I) = (\mathbf{W}_R\mathbf{X}_R - \mathbf{W}_I\mathbf{X}_I) + j(\mathbf{W}_R\mathbf{X}_I + \mathbf{W}_I\mathbf{X}_R),
\]
where the cross-terms $\mathbf{W}_R\mathbf{X}_I$ and $\mathbf{W}_I\mathbf{X}_R$ explicitly couple amplitude and phase, allowing the model to learn joint scaling and rotation in the complex plane. To isolate the contribution of this principled structure, we introduce a \textit{real-valued} ablation that decomposes the operation into two independent real linear layers, discarding the cross-terms entirely:
\[
\hat{\mathbf{X}}_{\text{freq}}^{\text{(real)}} = \mathbf{W}_R^{\text{ind}}\mathbf{X}_R + j\mathbf{W}_I^{\text{ind}}\mathbf{X}_I.
\]
While both variants maintain identical parameter counts, the real mode severs the coupling between real and imaginary components, reducing the operation to independent channel-wise projections.

\begin{table}[h]
\centering
\caption{Frequency predictor design: complex-valued \textit{vs.} real-valued decomposition.}
\label{tab:freq_ablation}
\resizebox{\textwidth}{!}{
\begin{tabular}{cc|c|cc|c}
\toprule
Dataset & $K$ & Horizon & Complex (MSE / MAE) & Real (MSE / MAE) & $\Delta$MSE (\%) \\
\midrule
\multirow{4}{*}{Traffic} & \multirow{4}{*}{1} 
& 96  & 0.373 / 0.259 & 0.373 / 0.259 & +0.0 \\
& & 192 & 0.394 / 0.269 & 0.393 / 0.268 & --0.3 \\
& & 336 & 0.406 / 0.274 & 0.406 / 0.274 & +0.0 \\
& & 720 & 0.426 / 0.292 & 0.437 / 0.292 & +2.6 \\
\midrule
\multirow{4}{*}{Solar} & \multirow{4}{*}{1} 
& 96  & 0.186 / 0.236 & 0.190 / 0.237 & +2.2 \\
& & 192 & 0.201 / 0.248 & 0.209 / 0.261 & +4.0 \\
& & 336 & 0.207 / 0.254 & 0.214 / 0.267 & +3.4 \\
& & 720 & 0.211 / 0.258 & 0.216 / 0.268 & +2.4 \\
\midrule
\multirow{4}{*}{ETTh2} & \multirow{4}{*}{full} 
& 96  & 0.275 / 0.337 & 0.286 / 0.353 & +4.0 \\
& & 192 & 0.336 / 0.377 & 0.337 / 0.383 & +0.3 \\
& & 336 & 0.355 / 0.396 & 0.356 / 0.399 & +0.3 \\
& & 720 & 0.384 / 0.422 & 0.385 / 0.424 & +0.3 \\
\midrule
\multirow{4}{*}{Exchange} & \multirow{4}{*}{full} 
& 96  & 0.084 / 0.201 & 0.088 / 0.207 & +4.8 \\
& & 192 & 0.175 / 0.397 & 0.180 / 0.303 & +2.9 \\
& & 336 & 0.321 / 0.411 & 0.331 / 0.419 & +3.1 \\
& & 720 & 0.910 / 0.725 & 1.147 / 0.814 & +26.0 \\
\bottomrule
\end{tabular}
}
\end{table}

The results reveal a clear data- and horizon-dependent pattern, summarized in three regimes:

\paragraph{Hybrid periodic--transient regimes ($K$=1).} On Solar, where a single dominant harmonic coexists with non-stationary residuals, the complex predictor consistently outperforms the real variant across all horizons, with the MSE gap widening from 2.2\% at $H$=96 to 4.0\% at $H$=192. This validates our hypothesis that decoupled real/imaginary mappings corrupt phase consistency: even minor phase distortions in the Frequency Branch propagate into the Time Branch during final aggregation, compounding the overall prediction error. The complex cross-terms act as a structural regularizer that anchors the principal harmonic to its true phase, which is essential when dual-branch collaboration is active.

\paragraph{Long-horizon spectral extrapolation.} In the pure frequency-domain regime, the complex predictor demonstrates critical robustness at long horizons. On Exchange, both variants perform comparably for $H$$\le$336; however, at $H$=720 the real-valued decomposition suffers catastrophic degradation (MSE 1.147 \textit{vs.} 0.910, +26.0\%). We attribute this to accumulated phase drift: without cross-term constraints, independent real/imaginary projections gradually lose global phase alignment during spectral extrapolation, and this error is amplified over long prediction windows. The complex formulation, by contrast, inherently preserves the rotational structure of Fourier coefficients, preventing such drift.

\paragraph{Short-horizon and stable-periodic tasks.} On ETTh2 and Traffic, the performance gap remains marginal ($<$4\%). These datasets either exhibit stable multi-frequency periodicity (ETTh2) or strong deterministic cycles with limited non-stationary residuals (Traffic), where the forecasting task is less sensitive to fine-grained phase constraints. Nevertheless, the complex predictor achieves strictly lower average MSE on both benchmarks, confirming that amplitude-phase coupling provides a universally beneficial inductive bias at no additional parameter cost.

In summary, the complex-valued linear predictor is not merely an implementation detail, but a principled design choice that safeguards phase consistency. Its advantages become pronounced in hybrid routing scenarios and long-horizon forecasting, where phase errors would otherwise propagate or accumulate.

\section{Robustness Analysis}
\label{sec:robustness}

To further validate the practical applicability of RouteTS, we evaluate its performance under two common real-world perturbations: high-frequency noise injection and varying input lengths. We select representative baselines from each paradigm: PhaseFormer (Transformer-based), DLinear (Linear-based), and FBM (Frequency-based).

\subsection{Gaussian Noise Injection}
\label{sec:robustness_noise}

In real-world scenarios, sensor-collected time series are inevitably contaminated by high-frequency random noise. We inject Gaussian white noise with increasing standard deviation ratios ($\sigma_{\text{noise}} \in \{0.0, 0.2, 0.4, 0.6, 0.8\}$) into the input sequences and evaluate model resilience on Solar ($K=1$) and ETTm2 ($K=\text{full}$), which represent single-frequency and complex periodic regimes, respectively.

\begin{figure}[t]
\centering
\includegraphics{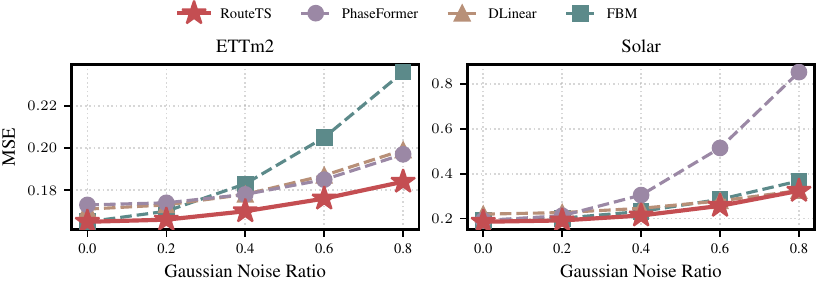}
\caption{Robustness evaluation under varying Gaussian noise ratios on Solar ($K=1$) and ETTm2 ($K=\text{full}$). RouteTS demonstrates superior noise immunity and prevents the catastrophic failure observed in attention-based models by effectively functioning as an adaptive low-pass filter.}
\label{fig:noise_robustness}
\end{figure}

As illustrated in Figure~\ref{fig:noise_robustness}, RouteTS demonstrates superior noise immunity across all noise levels. On Solar, the attention-based PhaseFormer exhibits drastic performance degradation as noise increases, with MSE rising by over $340\%$ (from $0.192$ to $0.853$). In contrast, RouteTS degrades gracefully from $0.186$ to $0.324$, effectively functioning as an adaptive low-pass filter that selectively extracts dominant periodicities while isolating high-frequency noise. On ETTm2, RouteTS maintains consistent dominance across the entire noise spectrum, while FBM suffers noticeable degradation at higher noise ratios ($0.236$ at $\sigma=0.8$ vs. $0.184$ for RouteTS).

\subsection{Varying Input Length}
\label{sec:robustness_lookback}

\begin{figure}[t]
\centering
\includegraphics{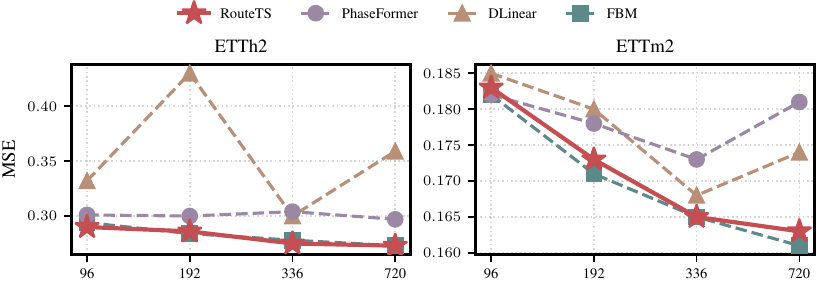}
\caption{Robustness of different models against varying look-back window lengths on ETTh2 and ETTm2.}
\label{fig:lookback_robustness}
\end{figure}

We further assess model stability against varying look-back window lengths ($L \in \{96, 192, 336, 720\}$) on ETTh2 and ETTm2. As shown in Figure~\ref{fig:lookback_robustness}, RouteTS exhibits monotonic performance improvement as the historical context expands, leveraging longer input to better capture multi-scale periodic structures. On ETTh2, RouteTS achieves comparable performance to FBM ($0.273$ at $L=720$), while both maintain stable improvement as the look-back window extends. In contrast, DLinear exhibits severe non-monotonic behavior, with MSE spiking to $0.430$ at $L=192$ before recovering. On ETTm2, FBM achieves marginally lower MSE at $L=720$ ($0.161$ vs. $0.163$), yet RouteTS reaches nearly identical accuracy while using orders of magnitude fewer parameters ($6.21$MB vs. $907.82$MB, as shown in Figure~\ref{fig:efficiency_comparison}). This verifies that RouteTS efficiently exploits extended temporal dependencies without overfitting to spurious local patterns.

\end{document}